\documentclass{article}

\usepackage[dblblindworkshop, final]{neurips_2026}

\usepackage[utf8]{inputenc} 
\usepackage[T1]{fontenc}    
\usepackage{url}            
\usepackage{booktabs}       
\usepackage{amsfonts}       
\usepackage{amssymb}
\usepackage{nicefrac}       
\usepackage{microtype}      
\usepackage{xcolor}         
\usepackage{caption} 
\usepackage{enumitem} 
\usepackage{amsmath} 
\usepackage{todonotes} 
\usepackage{soul} 
\usepackage{graphicx} 
\usepackage{wrapfig} 
\usepackage{siunitx} 
\definecolor{mydarkblue}{rgb}{0,0.08,0.45}
\usepackage[ 
    colorlinks=true,
    linkcolor=mydarkblue,
    citecolor=mydarkblue,
    urlcolor=mydarkblue,
    filecolor=mydarkblue,
]{hyperref}
\usepackage{algorithm}
\usepackage{algorithmic}
\usepackage{subcaption}
\usepackage{float}
\usepackage{lipsum}

\title{How Geometry Shapes Generative Model Evaluation on Riemannian Manifolds}
\workshoptitle{Geometric Distributional Deep Learning}

\author{%
Misha Oberski$^{1}$\thanks{Equal contribution, correspondence to: \texttt{\{misha.oberski,\,valentijn.oldenburg\}@student.uva.nl}}, \;
Valentijn Oldenburg$^1$\footnotemark[1], \;
Xanthos Kekkos$^1$, \;
Soham Chaudhuri$^1$, \\
\textbf{Emma Oosterhuis}$^1$, \;
\textbf{Floris de Kam}$^1$, \;
\textbf{Sagi Schwartz}$^1$, \;
\textbf{Alison Pouplin}$^2$, \;
\textbf{Federica Valeau}$^{3}$ \\
$^1$University of Amsterdam, \;
$^2$Independent Researcher, \;
$^3$AMLab, University of Amsterdam
}

\begin{document}

\maketitle

\vspace{-0.8em}
\begin{abstract}
\vspace{-0.25em}
On Riemannian manifolds, generative-model evaluation is geometry-dependent since curvature and dimension jointly determine distances and volume growth. Our framework systematically compares geodesic, ambient, and tangent-space versions of Wasserstein and MMD across different geometries. Using controlled-noise perturbations and Riemannian Flow Matching samples, we study metric agreement, noise sensitivity, and their relation to geometry-induced volume scaling. We find that no metric is uniformly reliable: evaluation distance can affect both measured discrepancies and noise sensitivity, varying across curvature and dimension. Our analysis leads to practical recommendations for reporting evaluation distances, comparing metric types, and calibrating metrics to the geometry of the samples.
\end{abstract}
\vspace{-0.8em}
\section{Introduction}\label{sec:introduction}
\vspace{-0.25em}

A generative model is typically evaluated by measuring how close its generated samples are to the target distribution. Most evaluation metrics are built on a distance between pairs of points. The Wasserstein distance uses it as a transport cost \citep{peyre2019computational,genevay2018learning,feydy2019interpolating}, and the maximum mean discrepancy (MMD) uses it as the argument of a kernel \citep{gretton2006kernel}.
In Euclidean space, there is only one natural way to measure this distance. On a Riemannian manifold, there are three: the geodesic distance on the manifold, the Euclidean distance between the ambient coordinates, or the Euclidean distance between points mapped to a common tangent space.
These distances agree in flat space and locally near the tangent origin, but can diverge with curvature and as points move apart.
Thus, evaluation on a manifold is itself a geometric choice.

We study two properties of an evaluation metric. Given several sets of generated samples, each compared with its target, a metric orders those generated sets from best to worst. We say that two metrics \emph{agree} when they produce the same ranking. Different metrics measure different aspects of closeness and are not expected to agree, but when they do not, a conclusion drawn with one metric does not transfer to another. We also say that a metric is \emph{sensitive} to a perturbation when its value grows with the size of the perturbation. We measure sensitivity as a signal-to-noise ratio, and a high ratio correlates with a more informative metric as it can detect perturbation reliably.

In this paper, we study how curvature and dimension change the agreement and the sensitivity of evaluation metrics on Riemannian manifolds. We consider the Wasserstein distance and the MMD, each computed with the geodesic, ambient, and tangent-space distance, which gives six versions of a metric for the same samples. Agreement is measured on the output of Riemannian Flow Matching (RFM, \citet{chen2023flow}), and sensitivity on target samples, to which we progressively add noise to mimic generated samples. We find that the curvature and the dimension influence both the agreement and the sensitivity of those metrics. Within one metric, changing the distance changes the value but rarely the ranking. Across metrics, the rankings differ, and at high dimension the MMD loses its sensitivity while the Wasserstein distance keeps it. Our code is available on \href{https://github.com/MishaOberski/eval-riemannian-gen-modelling}{GitHub}.
\section{Preliminaries}\label{sec:preliminaries}

\textbf{Riemannian manifolds.}\quad
A Riemannian manifold is a smooth manifold $\mathcal{M}$ equipped with a Riemannian metric $g$, i.e.\ a smoothly varying inner product $g_p : T_p\mathcal{M} \times T_p\mathcal{M} \to \mathbb{R}$ on the tangent space $T_p\mathcal{M}$ at each point $p \in \mathcal{M}$ \citep{lee1997riemannian}.
Intuitively, the metric specifies how to measure infinitesimal lengths and angles locally on the manifold, thereby inducing curve lengths, geodesics, as locally shortest paths, and the geodesic distance $d_g(x,y)$ between points $x,y \in \mathcal{M}$.
With $\mathcal{M}^d_\kappa$, we denote a $d$-dimensional manifold with sectional curvature $\kappa \in \mathbb{R}$, metric $g_\kappa$, and geodesic distance $d_\kappa := d_{g_\kappa}$.

\textbf{Riemannian Flow Matching (RFM).}\quad
RFM \citep{chen2023flow} learns a time-dependent tangent vector field $v_t : \mathcal{M}^d_\kappa \rightarrow T\mathcal{M}^d_\kappa$, with $\boldsymbol{v}_t(x)\in T_x\mathcal{M}_\kappa^d$, that transports a base distribution $p_0$ to a target distribution $p_1$.
On geometries with closed-form geodesics (e.g., spherical or hyperbolic space), training is simulation-free: for samples $x_0\sim p_0$, $x_1\sim p_1$, and time $t \in [0,1]$, both (\emph{i}) the conditional path $\psi_t(x_0 \mid x_1)$ connecting $x_0$ to $x_1$, with point $x_t =\psi_t(x_0\mid x_1)$ at time $t$, and (\emph{ii}) the target velocity $\dot{\boldsymbol{x}}_t$, which specifies how the path moves at time $t$, can be computed in closed form.
The geodesic connecting $x_0$ and $x_1$ can be expressed in terms of exponential and logarithmic maps
\begin{equation}\label{eq:exp_log_maps}
    x_t=\psi_t^\kappa(x_0\mid x_1)=\exp_{x_1}^\kappa\!\!\big(\alpha(t)\log_{x_1}^\kappa(x_0)\big),\qquad \alpha(t)=1-t.
\end{equation}
\section{Experimental setup}\label{sec:cross-curvature_experimental_setup}

For our RFM and noise experiments, we consider constant-$\kappa$ manifolds, giving a controlled setting for cross-curvature comparison.
Positive $\kappa$ denotes spherical space $\mathbb{S}^d_\kappa$, $\kappa = 0$ Euclidean space $\mathbb{R}^d$, and negative $\kappa$ hyperbolic space $\mathbb{H}^d_\kappa$, represented using the Poincar\'e ball model $\mathbb{B}^d_\kappa$.

\vspace{-0.1em}
\subsection{Geometries}\label{sec:geometries}
\vspace{-0.1em}

To compare geometries while varying only curvature and dimension, we parameterise all spaces $\mathcal{M}_\kappa^d$ in geodesic polar coordinates around a fixed origin $o \in \mathcal{M}^d_\kappa$. Any point $x \in \mathcal{M}^d_\kappa$ is then described by its geodesic radius $r = d_\kappa(o, x)$ and an angular direction $\theta \in \mathbb{S}^{d-1}$, where $\mathbb{S}^{d-1}$ is a unit sphere of directions at $o$. 
In these coordinates, the Riemannian metric takes the form
  $ g_\kappa = dr^2 + \mathrm{sn}_\kappa(r)^2 \, d\Omega^2, $
where $d\Omega^2$ denotes the angular component, i.e.\ motion along directions around $o$, and
\begin{equation}\label{eq:S_k}
\operatorname{sn}_\kappa(r)=
\sin(\sqrt{\kappa}r)/\sqrt{\kappa}\text{ for }\kappa>0;\quad
r\text{ for }\kappa=0;\quad
\sinh(\sqrt{-\kappa}r)/\sqrt{-\kappa}\text{ for }\kappa<0,
\end{equation}
determines how angular directions are scaled at geodesic radius $r$; in constant curvature this follows from the Jacobi equation \citep{petersen2006riemannian, lang2020riemannian}.
These cases together form a unified parameterisation for constant-curvature geometry through $\kappa$.
(Appendix~\ref{ap:experimental_setup} lists used $\kappa$ and $d$ values.)

\textbf{Volume scaling.}\quad
To quantify how curvature changes local volume, we compare infinitesimal volume elements on the tangent space and manifold.
In geodesic polar coordinates around a fixed $o$, the volume element is $d\mathrm{vol}_\kappa = \mathrm{sn}_\kappa(r)^{d-1}\,dr\,d\Omega$, whereas in the flat tangent space, it is $r^{d-1}\, dr\, d\Omega$ \citep{gallot2004riemannian}.
Their ratio gives the Jacobian determinant of the exponential map \citep{lang2020riemannian}:
\begin{equation}\label{eq:Jacobian_factor}
    J_\kappa(r) := \left|\det D_{\boldsymbol{v}} \exp_o^\kappa\right| = \frac{\mathrm{sn}_\kappa(r)^{d-1}\, dr\, d\Omega}{r^{d-1}\,dr\,d\Omega} = \left(\frac{\mathrm{sn}_\kappa(r)}{r}\right)^{d-1},
\end{equation}
where $\boldsymbol{v} \in T_o\mathcal M_\kappa^d$ with $\|\boldsymbol{v}\|=r$, which measures the local volume scaling when mapped by $\exp_o^\kappa$ from $T_o\mathcal M_\kappa^d$ to $\mathcal{M}_\kappa^d$ at radius $r$. Given $p$ on $\mathcal{M}_\kappa^d$, we quantify the amount of volume scaling acting on the distribution, via the mean of the Jacobian at points $x \sim p$
\begin{equation}\label{eq:mean_volume_growth}
    \mu_{ J_\kappa}(p) := \mathbb{E}_{x\sim p}\!\left[ J_\kappa(r(x))\right].
\end{equation}

\vspace{-1em}
\subsection{Distributions}\label{sec:distributions}
\vspace{-0.1em}

We define distributions $\tilde p_0$ and $\tilde p_1$ on the tangent space $T_o \mathcal{M}^d_\kappa$ of fixed origin $o$.
We choose $o$ as the north pole on $\mathbb{S}^d_\kappa$, the Euclidean origin on $\mathbb{R}^d$, and the centre of the Poincaré ball on $\mathbb{H}^d_\kappa$.\footnote{These are coordinate conventions: by homogeneity of the constant-$\kappa$ spaces, the origin choice is arbitrary up to isometry.}
For $z_i\sim\tilde p_i$ and $i\in\{0,1\}$, all experiments map curvature-normalised tangent samples to the manifold as $x_i^0=z_i$ for $\kappa=0$ and $x_i^\kappa=\exp_o^\kappa(R_\kappa z_i)$ for $\kappa\neq0$, with $R_\kappa=|\kappa|^{-1/2}$.
This keeps distributions at comparable relative scale across curvatures, so evaluation differences are not merely due to the same tangent samples occupying different fractions of the manifold; Appendix~\ref{ap:tangent_space_normalisation} gives the derivation.
We use three families of tasks with $\tilde p_0=\mathcal N(\mathbf{0},0.3^2 \mathbf{I}_d)$ centered at $o$.
As targets $p_1$, we use:
\begin{itemize}[leftmargin = 20pt, topsep = -1pt, itemsep = -1pt]
    \item[(\emph{i})] An isotropic Gaussian, giving the unimodal Gaussian-to-Gaussian (GtG) baseline.
    \item[(\emph{ii})] Symmetric Gaussian Ring (SGR), a rotationally symmetric target with mass concentrated near radius $r$ (``shell-shaped'' for $d > 2$). Varying $r$ is designed to study radial volume growth.
    \item[(\emph{iii})] Mixture-of-Gaussians (MoG), a multimodal target placed symmetrically in $T_o\mathcal M$.\
\end{itemize}

\vspace{-0.15em}
\subsection{Metrics}\label{sec:metrics}
\vspace{-0.15em}

We write $d_{\mathrm{eval}}$ for the evaluation of distance metrics, instantiated (\emph{i}) intrinsically as geodesic distance $d_\mathrm{geodesic} := d_\kappa$ on $\mathcal M_\kappa^d$, (\emph{ii}) extrinsically as Euclidean distance $d_\mathrm{ambient}$ in ambient coordinates, or (\emph{iii}) locally as Euclidean distance $d_\mathrm{tangent}$ after log-mapping to $T_o\mathcal M_\kappa^d$ (details in Appendix~\ref{ap:metrics}).

\textbf{Wasserstein.}\quad 
To evaluate how well the generated distribution $\hat{p}_1$ matches the target $p_1$, we use the $1$-Wasserstein distance $W_1$ on $\mathcal{M}_\kappa^d$ (also known as the Earth Mover's distance \citep{andoni2008earth}):
\begin{equation}\label{eq:Wasserstein}
W_1(p_1, \hat{p}_1)
=
\inf_{\pi \in \Pi(p_1, \hat{p}_1)}
\int_{\mathcal{M}_\kappa^d \times \mathcal{M}_\kappa^d} d_\mathrm{eval} (y, \hat{y})\, d\pi(y, \hat{y}),
\end{equation}
where $\pi \in \Pi(p_1,\hat p_1)$ is a coupling over samples $y \sim p_1$ and $\hat{y} \sim \hat{p}_1$ within the set of all couplings $\Pi$ \citep{peyre2019computational}, giving the lowest-cost transport between $p_1$ and $\hat{p}_1$.
In practice, we approximate $W_1$ by entropically regularized Sinkhorn divergence \citep{feydy2019interpolating}; see Appendix~\ref{ap:metrics_accuracy}.

\textbf{MMD.}\quad
MMD compares distributions through average kernel similarities \citep{gretton2006kernel}. 
Using the same target and candidate samples $\{y_i\}_{i=1}^n$ and $\{\hat y_i\}_{i=1}^n$, we compute
\begin{equation}\label{eq:MMD_squared}
    \mathrm{MMD}^2_{d_{\mathrm{eval}}}
    =
    \frac{1}{n^2}\sum_{i,j} k(y_i,y_j)
    -\frac{2}{n^2}\sum_{i,j} k(y_i,\hat y_j)
    +\frac{1}{n^2}\sum_{i,j} k(\hat y_i,\hat y_j),
\end{equation}
with RBF kernel $k(x,x')=\exp{(-d_{\mathrm{eval}}(x,x')^2/(2\ell^2))}$ and $\ell$ as median non-zero pairwise distance.

\textbf{SNR.}\quad
To separate metric sensitivity from model error, we evaluate each metric on controlled perturbations of the target distribution. 
For a metric $D$, let $D_s(\sigma)$ denote the metric between the clean and perturbed target distributions in resampling replicate $s$.
Given target samples $z \sim p_1$ on the tangent space, we construct noisy candidates by adding Gaussian noise: we sample $u_{s,\sigma}\sim\mathcal{N}_{T_z\mathcal{M}_\kappa^d}(\mathbf{0},\frac{\sigma^2}{d}\mathbf{I}_d)$ and then project $x_{s,\sigma}=\exp_x^\kappa(u_{s,\sigma})$.
For $L$ ordered noise levels ($\sigma_0$ up to $\sigma_{L-1}$), we define the signal-to-noise ratio between levels $\sigma_i<\sigma_j$ as
\begin{wrapfigure}{r}{0.45\textwidth}
    \vspace{-1.2em}
    \centering
    \captionsetup{
        font=normalsize,
        labelfont=normalfont,
        textfont=normalfont,
        skip=2pt,
        justification=raggedright,
        singlelinecheck=false
    }
    \includegraphics[width=0.98\linewidth]{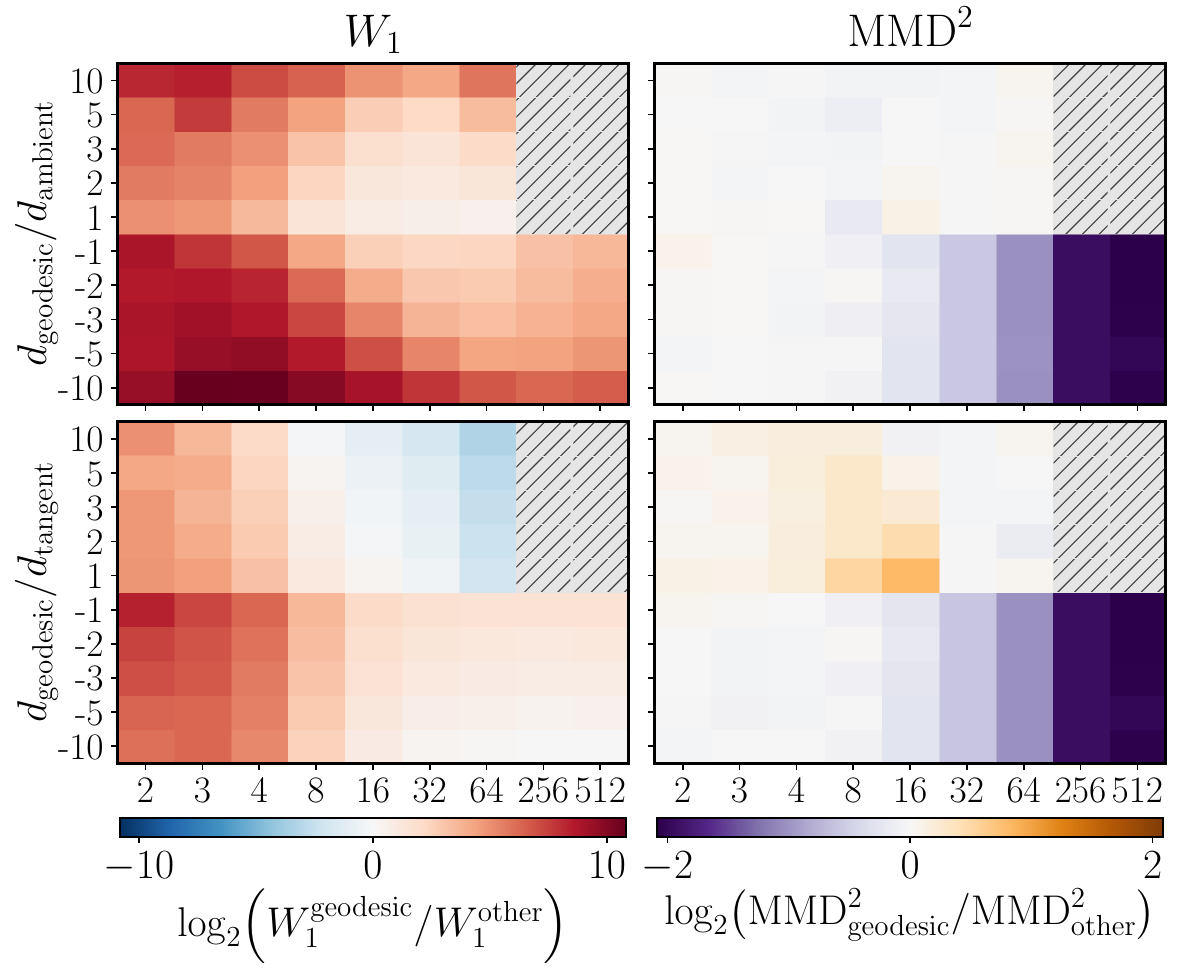}
    \vspace{-0.1em}
    \caption{
    \textbf{Metric discrepancies depend on evaluation geometry.}
    Cells show $\log_2(D_{\mathrm{geodesic}}/D_{\mathrm{other}})$ over $(\kappa,d)$ for GtG, comparing geodesic evaluation with ambient (top) and tangent-space (bottom) distances.
    Positive values mean geodesic eval.\ yields a larger discrepancy.
    $W_1$ is mostly larger under geodesic, whereas $\mathrm{MMD}$ decreases in high $d$.
    }
    \label{fig:geometric_metric_ratios_gtg}
    \vspace{-4em}
\end{wrapfigure}\vspace{-0.7em}
\begin{equation}\label{eq:metric_snr}
    \mathrm{SNR}_D(i,j)
    :=
    \frac{
    \mathbb E_s[D_s(\sigma_j)-D_s(\sigma_i)]
    }{
    \mathrm{Std}_s[D_s(\sigma_j)-D_s(\sigma_i)]
    }.
\end{equation}
A high SNR means the metric detects degradation above sampling variability.
Averaging these values gives a single summary value for each noise-response curve, which we call the mean consecutive SNR:  $\overline{\mathrm{SNR}}_D=\frac{1}{L-1}\sum_{i=0}^{L-2}\mathrm{SNR}_D(i,i+1)$.
Finally, we apply the metrics to RFM samples, where errors arise from learned transport instead of controlled noise.
\section{Experiments and analysis}\label{sec:experiments}

We discuss the main findings in this section; further results are provided in Appendix~\ref{ap:additional_results}.

\begin{wrapfigure}[39]{r}{0.49\textwidth} 
    \vspace{-1em}
    \centering
    \captionsetup{
        font=normalfont,
        labelfont=normalfont,
        textfont=normalfont,
        skip=2pt,
        justification=justified,
        singlelinecheck=false
    }
    \includegraphics[width=0.98\linewidth]{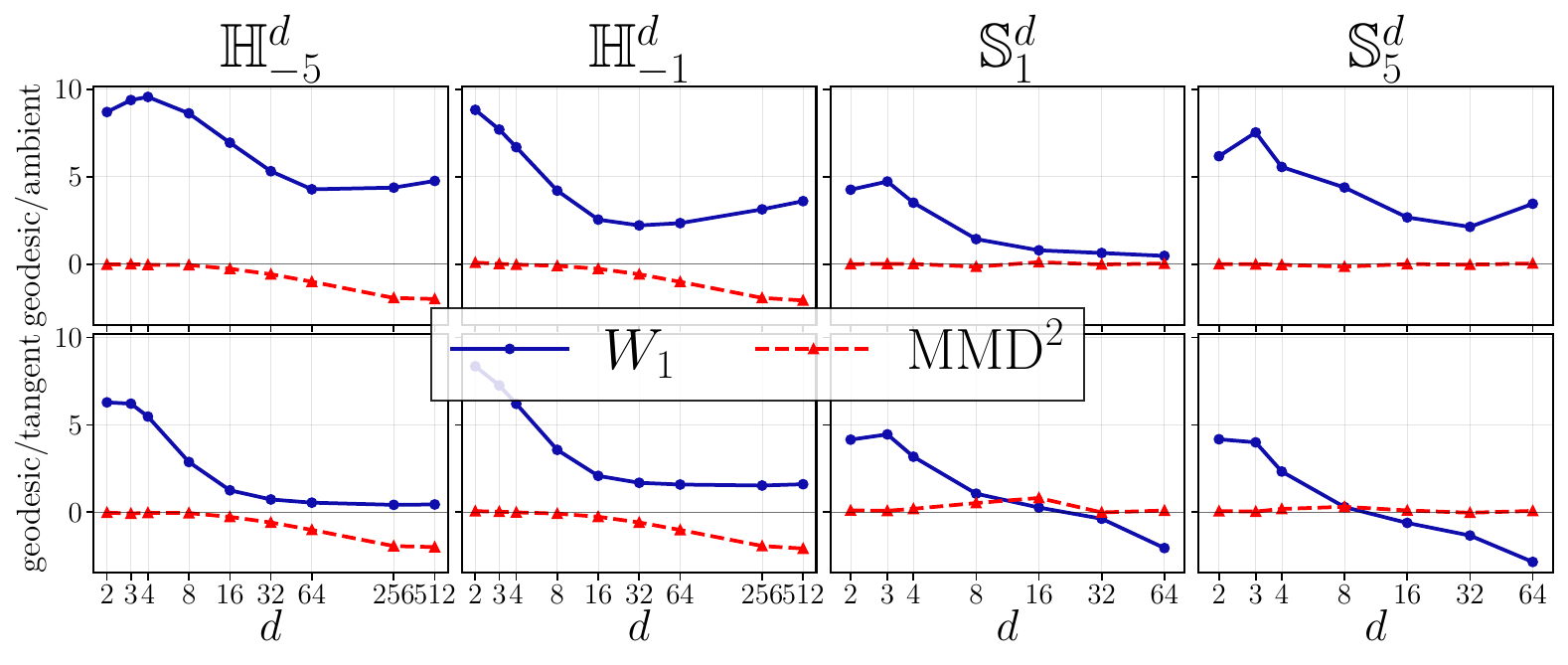}
    \caption{
    \textbf{Evaluation distance changes metric trends.}
    Curves show $\log_2(D_{\mathrm{geodesic}}/D_{\mathrm{other}})$ across dimensions for four curvatures.
    $W_1$ and $\mathrm{MMD}^2$ respond differently to the same change from intrinsic to non-intrinsic evaluation.}
    \label{fig:intrinsic_vs_nonintrinsic_curves}
    \vspace{0.1em}
    \includegraphics[width=0.98\linewidth]{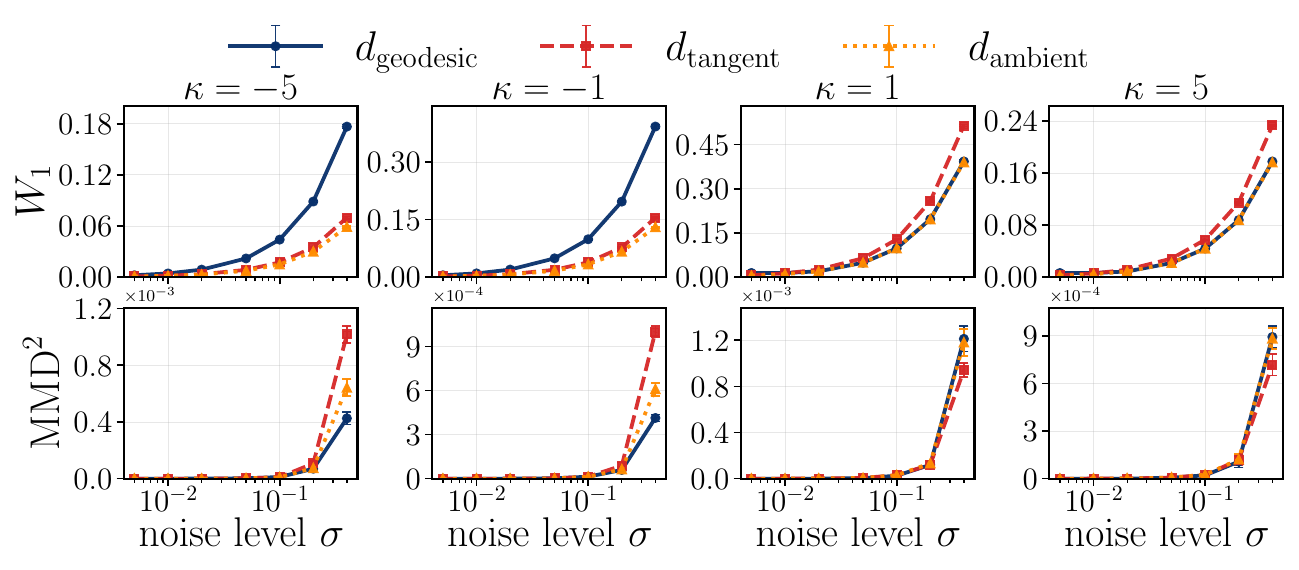}
    \caption{\textbf{Metrics detect sample worsening differently across geometries.}
    At $d=16$, $W_1$ and $\mathrm{MMD}^2$ generally increase with controlled noise, but the response differs across curvature and across evaluation distance.}
    \label{fig:raw_noise_response_d16}
    \vspace{0.1em}
    \includegraphics[width=0.98\linewidth]{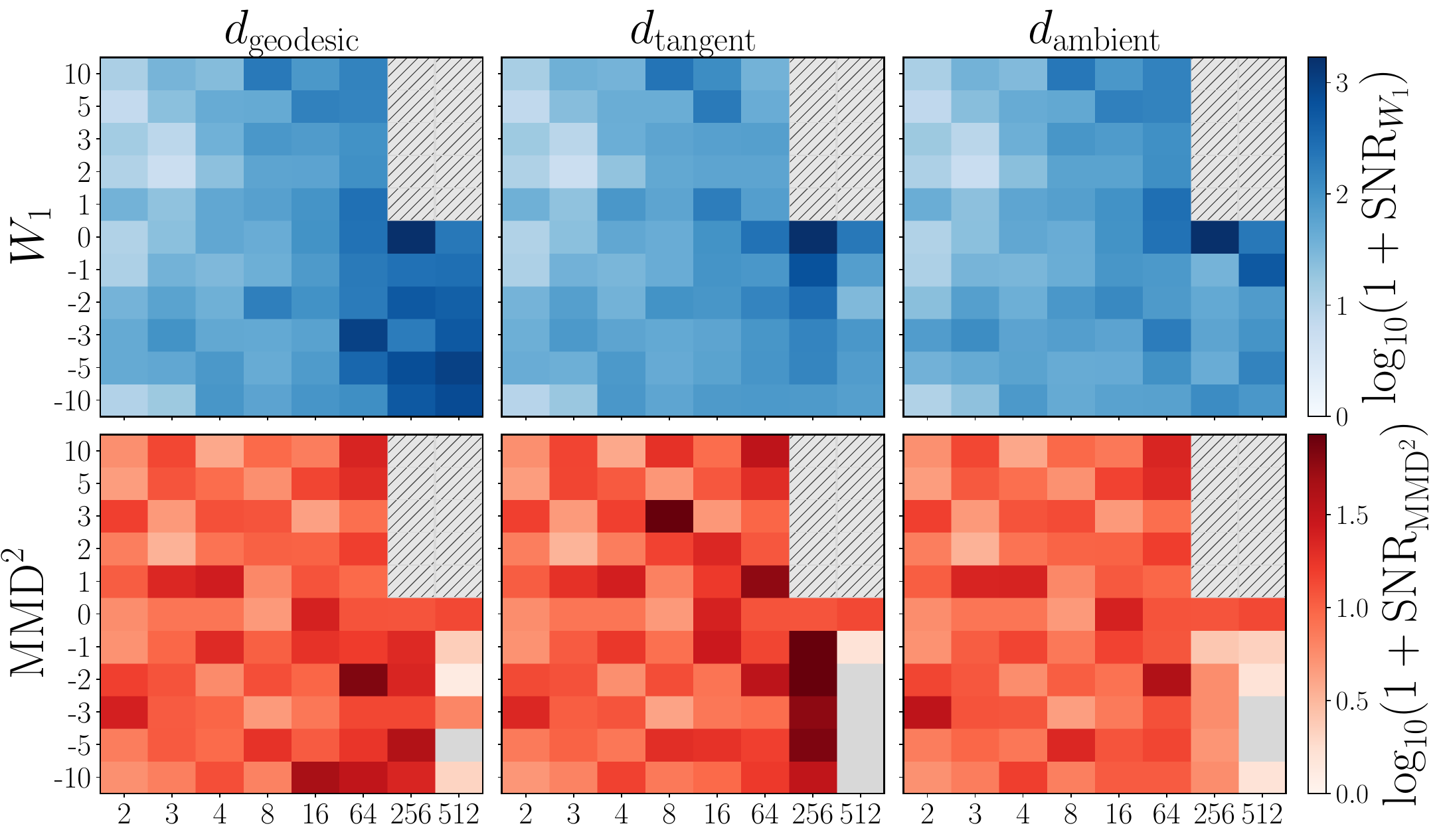}
    \caption{\textbf{Noise-detection reliability is geometry-dependent.}
    Cells show global SNR $\mathrm{SNR}_D(0,L-1)$ over $(\kappa,d)$, summarising how reliably each metric detects the controlled perturbations.
    Reliability varies with both evaluation distance and curvature.}
    \label{fig:global_snr_heatmap}
    \vspace{-0.1em}
\end{wrapfigure}

\textbf{Metric dependence.}\quad
Figure~\ref{fig:geometric_metric_ratios_gtg} shows contrasting dimensional trends in hyperbolic space: the gap between geodesic and non-intrinsic $W_1$ decreases with $d$, whereas geodesic $\mathrm{MMD}^2$ increasingly diverges from both tangent and ambient.
Figure~\ref{fig:intrinsic_vs_nonintrinsic_curves} shows the same trend as dimensionality increases: spherical spaces show weaker divergence, with tangent and ambient approximations behaving similarly. Hence, dimension does not uniformly amplify geometric error: it changes evaluation in a metric- and curvature-dependent manner. 
Appendix~\ref{ap:additional_results} shows a complementary pattern: rankings are relatively stable across evaluation distances within a metric type, but weak and target-dependent across families (Figure~\ref{fig:metric_agreement_all_app}). 

\textbf{Controlled noise perturbations.} \quad
Figure~\ref{fig:raw_noise_response_d16} shows that $W_1$ and $\mathrm{MMD}^2$ generally increase monotonically with noise, but with curvature- and distance-dependent response strength, especially in negatively curved spaces.
Figure~\ref{fig:global_snr_heatmap} extends this across the sweep: global SNR (computed as $\mathrm{SNR}_D(0,L-1)$ from Eq.~\eqref{eq:metric_snr}) varies systematically over $(\kappa,d)$, showing that reliability is not a property of the metric alone.
Appendix~\ref{ap:controlled_noise_perturbations} confirms this pattern with a monotonicity check: $W_1$ remains near-monotone across dimensions, whereas $\mathrm{MMD}^2$ weakens at high $d$ (Table~\ref{tab:rq1_det_scores}).
Thus, distance representation affects not only the reported discrepancy size, but also how reliably sample-quality degradations are detected.

\textbf{Geometric interpretation.}\quad
Curvature changes local volume through $J_\kappa$ (Eq.~\eqref{eq:Jacobian_factor}), so the same intrinsic perturbation can yield different pairwise distances under geodesic, tangent-space, or ambient evaluation.
This helps explain $W_1$: where volume grows rapidly, ambient distances can underestimate intrinsic separations, potentially making geodesic evaluation more sensitive to noise.
Table~\ref{tab:volume_snr_correlation} supports this for ambient $W_1$, while the weaker tangent-$W_1$ correlation suggests that log-mapping preserves enough radial structure for volume scaling alone not to predict the gap.
For $\mathrm{MMD}^2$, Eq.~\eqref{eq:MMD_squared} averages RBF similarities using $d_{\mathrm{eval}}$ and a median-distance bandwidth, so curvature-induced distance shifts alter both kernel values and bandwidths, yielding positive associations for the ambient and tangent alternatives.
Such kernel discrepancies are also known to lose sensitivity in high dimensions \citep{reddi2014decreasingpowerkerneldistance}.
Thus, volume scaling helps explain the geometry-dependent reliability gaps, but the different $W_1$ and $\mathrm{MMD}^2$ patterns show that metric-specific computation also matters.
\vspace{-0.25em}
\section{Conclusion}\label{sec:conclusion}
\vspace{-0.15em}
\begin{table}[b]
\centering
\vspace{-2em}
\begin{minipage}[t]{0.33\linewidth}
\centering
\captionsetup{
    type=table,
    font=normalsize,
    labelfont=normalfont,
    textfont=normalfont,
    skip=2pt,
    justification=raggedright,
    singlelinecheck=false
}
\caption{
Spearman correlations between $\log\mu_{J_\kappa}(p_1)$ and
$\log_2(\overline{\mathrm{SNR}}_{\mathrm{geodesic}}/
\overline{\mathrm{SNR}}_{\mathrm{other}})$ on GtG.
Positive values mean $d_{\mathrm{geodesic}}$ gains reliability over $d_{\mathrm{other}}$.
Full pattern in Figure~\ref{fig:volume_snr_scatter_gtg}.
}
\renewcommand{\arraystretch}{0.85}
\vspace{-0.2em}
\footnotesize
\begin{tabular}{@{}lcc@{}}
\toprule
 & $d_{\mathrm{ambient}}$ & $d_{\mathrm{tangent}}$ \\
\midrule
$W_1$ & 0.63 & 0.20 \\
$\mathrm{MMD}^2$ & 0.62 & 0.62 \\
\bottomrule
\end{tabular}
\label{tab:volume_snr_correlation}
\end{minipage}
\hfill
\begin{minipage}[t]{0.66\linewidth}
\vspace{0.1em}
\centering
\captionsetup{
    type=table,
    font=normalsize,
    labelfont=normalfont,
    textfont=normalfont,
    skip=2pt,
    justification=raggedright,
    singlelinecheck=false
}
\setlength{\tabcolsep}{2pt}
\renewcommand{\arraystretch}{0.86}
\caption{Practical guidance for manifold sample evaluation.}
\vspace{-0.2em}
\label{tab:practical_guidance}
\footnotesize
\begin{tabular}{@{}p{0.20\linewidth}p{0.56\linewidth}p{0.16\linewidth}@{}}
\toprule
Situation & Recommendation & Support \\
\midrule
All evaluations
& Report $d_{\mathrm{eval}}$ as part of the metric; for MMD, also state the kernel bandwidth rule.
& Figures~\ref{fig:geometric_metric_ratios_gtg}, \ref{fig:intrinsic_vs_nonintrinsic_curves}, \ref{fig:raw_noise_response_d16}, \ref{fig:global_snr_heatmap} \\
High-$d$, $\kappa<0$
& Avoid ambient-only conclusions; verify important results with $d_\mathrm{geodesic}$ or $d_\mathrm{tangent}$.
& Figure~\ref{fig:geometric_metric_ratios_gtg} and Table~\ref{tab:volume_snr_correlation} \\
Comparison of models
& Use at least two metric types, since rankings need not transfer between e.g.\ $W_1$ \& $\mathrm{MMD}^2$.
& Figure~\ref{fig:metric_agreement_all_app} \\
High-$d$ MMD
& Run a small controlled-noise check before relying on MMD as the main quality signal
& Table~\ref{tab:rq1_det_scores}, Figures~\ref{fig:global_snr_heatmap}, \ref{fig:absolute_snr_vs_dimension_app} \\
\bottomrule
\end{tabular}
\end{minipage}
\vspace{-0.8em}
\end{table}
Our study shows that neither $W_1$ nor $\mathrm{MMD}^2$ is uniformly reliable across geometries.
Where \citet{kansal2023evaluating} show that generative-model metrics can have different sensitivities in Euclidean data, we show that this non-interchangeability becomes geometric on manifolds: both metric family and evaluation distance can change measured discrepancies and noise sensitivity.
At the metric type level, agreement is weak and target-dependent; at the distance level, geodesic, ambient, and tangent-space evaluations can change the reported discrepancy even for the same metric.
In particular, $\mathrm{MMD}^2$ loses sensitivity in some high-$d$ settings, whereas $W_1$ remains comparatively consistent.
The dependence on curvature and dimension is consistent with curvature-induced volume scaling, suggesting that intrinsic evaluation is especially important in high-$d$ hyperbolic spaces.
Thus, we recommend to report the evaluation distance, to compare different metric types, and to check their sensitivity to noise, especially in these high-$d$ hyperbolic spaces.
Future work can use these diagnostics to guide metric selection and representation design. In particular, extending the controlled evaluation framework to learned latent manifolds \citep{yue2026imagegenerationsphereencoder, bu2026ggballgraphgenerativemodel} would allow the latent geometry and its evaluation metric to be selected jointly, potentially leading to a more sensitive representation.
Table~\ref{tab:practical_guidance} summarises what we recommend checking when evaluating models on manifolds.
\section*{Acknowledgements}
We kindly thank Maria Esteban Casadevall for helpful discussions and insights during the early stages of this work.
We also thank SURF (\url{www.surf.nl}) for the support in using the National Supercomputer Snellius.

\bibliographystyle{unsrtnat}
\bibliography{references}

@book{petersen2006riemannian,
  title={Riemannian geometry},
  author={Petersen, Peter},
  year={2006},
  publisher={Springer}
}

@misc{chen2023flow,
      title={Flow Matching on General Geometries}, 
      author={Ricky T. Q. Chen and Yaron Lipman},
      year={2024},
      eprint={2302.03660},
      archivePrefix={arXiv},
      primaryClass={cs.LG},
      url={https://arxiv.org/abs/2302.03660}, 
}

@misc{karcher2014riemannian,
      title={{Riemannian} Center of Mass and so called {Karcher} mean}, 
      author={Hermann Karcher},
      year={2014},
      eprint={1407.2087},
      archivePrefix={arXiv},
      primaryClass={math.HO},
      url={https://arxiv.org/abs/1407.2087}, 
}

@inproceedings{bachmann2020constant,
author = {Bachmann, Gregor and B{\'e}cigneul, Gary and Ganea, Octavian-Eugen},
title = {Constant curvature graph convolutional networks},
year = {2020},
publisher = {JMLR.org},
booktitle = {Proceedings of the 37th International Conference on Machine Learning},
articleno = {46},
numpages = {11},
series = {ICML'20}
}

@article{afsari2011riemannian,
  title={Riemannian {$L^{p}$} center of mass: existence, uniqueness, and convexity},
  author={Afsari, Bijan},
  journal={Proceedings of the American Mathematical Society},
  volume={139},
  number={2},
  pages={655--673},
  year={2011}
}

@inproceedings{andoni2008earth,
author = {Andoni, Alexandr and Indyk, Piotr and Krauthgamer, Robert},
title = {Earth mover distance over high-dimensional spaces},
year = {2008},
publisher = {Society for Industrial and Applied Mathematics},
address = {USA},
booktitle = {Proceedings of the Nineteenth Annual ACM-SIAM Symposium on Discrete Algorithms},
pages = {343–352},
numpages = {10},
location = {San Francisco, California},
series = {SODA '08}
}

@misc{feydy2019interpolating,
      title={Interpolating between Optimal Transport and {MMD} using {Sinkhorn} Divergences}, 
      author={Jean Feydy and Thibault Séjourné and François-Xavier Vialard and Shun-ichi Amari and Alain Trouvé and Gabriel Peyré},
      year={2018},
      eprint={1810.08278},
      archivePrefix={arXiv},
      primaryClass={math.ST},
      url={https://arxiv.org/abs/1810.08278}, 
}

@misc{lang2020riemannian,
  author       = {Urs Lang},
  title        = {Lecture Notes on {Riemannian} Geometry},
  institution  = {ETH Z{\"u}rich},
  year         = {2020},
  note         = {Spring Semester 2020},
  url          = {https://metaphor.ethz.ch/x/2020/fs/401-3532-08L/sc/DG2_16June2020.pdf}
}

@book{gallot2004riemannian,
  title={Riemannian Geometry},
  author={Gallot, S. and Hulin, D. and Lafontaine, J.},
  isbn={9783540204930},
  lccn={04106886},
  series={Universitext},
  url={https://books.google.nl/books?id=6F4Umpws_gUC},
  year={2004},
  publisher={Springer Berlin Heidelberg}
}

@inproceedings{
valeau2026hyperbolic,
title={Hyperbolic Curvature as an Inductive Bias for Latent Space Flow Matching},
author={Federica Valeau and Maria Esteban-Casadevall and Erik J Bekkers},
booktitle={ICLR 2026 Workshop on Geometry-grounded Representation Learning and Generative Modeling},
year={2026},
url={https://openreview.net/forum?id=baSHbZuG5N}
}

@misc{yue2026imagegenerationsphereencoder,
      title={Image Generation with a Sphere Encoder}, 
      author={Kaiyu Yue and Menglin Jia and Ji Hou and Tom Goldstein},
      year={2026},
      eprint={2602.15030},
      archivePrefix={arXiv},
      primaryClass={cs.CV},
      url={https://arxiv.org/abs/2602.15030}, 
}

@misc{lipman2022flow,
      title={Flow Matching for Generative Modeling}, 
      author={Yaron Lipman and Ricky T. Q. Chen and Heli Ben-Hamu and Maximilian Nickel and Matt Le},
      year={2023},
      eprint={2210.02747},
      archivePrefix={arXiv},
      primaryClass={cs.LG},
      url={https://arxiv.org/abs/2210.02747}, 
}

@inproceedings{chen2018,
 author = {Chen, Ricky T. Q. and Rubanova, Yulia and Bettencourt, Jesse and Duvenaud, David K},
 booktitle = {Advances in Neural Information Processing Systems},
 editor = {S. Bengio and H. Wallach and H. Larochelle and K. Grauman and N. Cesa-Bianchi and R. Garnett},
 pages = {},
 publisher = {Curran Associates, Inc.},
 title = {Neural Ordinary Differential Equations},
 url = {https://proceedings.neurips.cc/paper_files/paper/2018/file/69386f6bb1dfed68692a24c8686939b9-Paper.pdf},
 volume = {31},
 year = {2018}
}

@misc{mathieu2020riemannian,
      title={Riemannian Continuous Normalizing Flows}, 
      author={Emile Mathieu and Maximilian Nickel},
      year={2020},
      eprint={2006.10605},
      archivePrefix={arXiv},
      primaryClass={stat.ML},
      url={https://arxiv.org/abs/2006.10605}, 
}

@misc{zaghen2025riemannian,
      title={Riemannian Variational Flow Matching for Material and Protein Design}, 
      author={Olga Zaghen and Floor Eijkelboom and Alison Pouplin and Cong Liu and Max Welling and Jan-Willem van de Meent and Erik J. Bekkers},
      year={2026},
      eprint={2502.12981},
      archivePrefix={arXiv},
      primaryClass={cs.LG},
      url={https://arxiv.org/abs/2502.12981}, 
}

@book{lee1997riemannian,
  author    = {John M. Lee},
  title     = {Riemannian Manifolds: An Introduction to Curvature},
  series    = {Graduate Texts in Mathematics},
  volume    = {176},
  publisher = {Springer-Verlag},
  address   = {New York},
  year      = {1997},
  isbn      = {978-0-387-22726-1},
  doi       = {10.1007/0-387-22726-1}
}

@misc{nowtricity2026co2,
  author = {Nowtricity},
  title  = {{CO$_2$} emissions per {kWh} in the {Netherlands}},
  year   = {2026},
  month  = May
}

@book{peyre2019computational,
      title={Computational Optimal Transport}, 
      author={Gabriel Peyré and Marco Cuturi},
      year={2020},
      eprint={1803.00567},
      archivePrefix={arXiv},
      primaryClass={stat.ML},
      url={https://arxiv.org/abs/1803.00567}, 
}

@misc{batle2014thomsonproblem,
title = {Generalized {Thomson} problem in arbitrary dimensions and {non-Euclidean} geometries},
journal = {Physica A: Statistical Mechanics and its Applications},
volume = {451},
pages = {237-250},
year = {2016},
issn = {0378-4371},
doi = {https://doi.org/10.1016/j.physa.2016.01.069},
url = {https://www.sciencedirect.com/science/article/pii/S0378437116001230},
author = {J. Batle and Armen Bagdasaryan and M. Abdel-Aty and S. Abdalla}
}

@misc{genevay2018learning,
      title={Learning Generative Models with {Sinkhorn} Divergences}, 
      author={Aude Genevay and Gabriel Peyré and Marco Cuturi},
      year={2017},
      eprint={1706.00292},
      archivePrefix={arXiv},
      primaryClass={stat.ML},
      url={https://arxiv.org/abs/1706.00292}, 
}

@misc{gretton2006kernel,
      title={A Kernel Method for the Two-Sample Problem}, 
      author={Arthur Gretton and Karsten Borgwardt and Malte J. Rasch and Bernhard Scholkopf and Alexander J. Smola},
      year={2008},
      eprint={0805.2368},
      archivePrefix={arXiv},
      primaryClass={cs.LG},
      url={https://arxiv.org/abs/0805.2368}, 
}

@misc{geoopt2020kochurov,
      title={Geoopt: {Riemannian} Optimization in {PyTorch}}, 
      author={Max Kochurov and Rasul Karimov and Serge Kozlukov},
      year={2020},
      eprint={2005.02819},
      archivePrefix={arXiv},
      primaryClass={cs.CG},
      url={https://arxiv.org/abs/2005.02819}, 
}

@misc{loshchilov2017decoupled,
      title={Decoupled Weight Decay Regularization}, 
      author={Ilya Loshchilov and Frank Hutter},
      year={2019},
      eprint={1711.05101},
      archivePrefix={arXiv},
      primaryClass={cs.LG},
      url={https://arxiv.org/abs/1711.05101}, 
}

@misc{you2026constant,
      title={Constant Metric Scaling in {Riemannian} Computation}, 
      author={Kisung You},
      year={2026},
      eprint={2601.10992},
      archivePrefix={arXiv},
      primaryClass={cs.LG},
      url={https://arxiv.org/abs/2601.10992}, 
}

@misc{ganea2018hyperbolic,
      title={Hyperbolic Neural Networks}, 
      author={Octavian-Eugen Ganea and Gary Bécigneul and Thomas Hofmann},
      year={2018},
      eprint={1805.09112},
      archivePrefix={arXiv},
      primaryClass={cs.LG},
      url={https://arxiv.org/abs/1805.09112}, 
}

@misc{van2023poincare,
      title={Poincar\'e {ResNet}}, 
      author={Max van Spengler and Erwin Berkhout and Pascal Mettes},
      year={2023},
      eprint={2303.14027},
      archivePrefix={arXiv},
      primaryClass={cs.CV},
      url={https://arxiv.org/abs/2303.14027}, 
}

@misc{de2022riemannian,
      title={Riemannian Score-Based Generative Modelling}, 
      author={Valentin De Bortoli and Emile Mathieu and Michael Hutchinson and James Thornton and Yee Whye Teh and Arnaud Doucet},
      year={2022},
      eprint={2202.02763},
      archivePrefix={arXiv},
      primaryClass={cs.LG},
      url={https://arxiv.org/abs/2202.02763}, 
}

@misc{reddi2014decreasingpowerkerneldistance,
      title={On the Decreasing Power of Kernel and Distance based Nonparametric Hypothesis Tests in High Dimensions}, 
      author={Sashank J. Reddi and Aaditya Ramdas and Barnabás Póczos and Aarti Singh and Larry Wasserman},
      year={2014},
      eprint={1406.2083},
      archivePrefix={arXiv},
      primaryClass={stat.ML},
      url={https://arxiv.org/abs/1406.2083}, 
}

@misc{jagvaral2024unified,
      title={Unified framework for diffusion generative models in {SO(3)}: applications in computer vision and astrophysics}, 
      author={Yesukhei Jagvaral and Francois Lanusse and Rachel Mandelbaum},
      year={2023},
      eprint={2312.11707},
      archivePrefix={arXiv},
      primaryClass={cs.LG},
      url={https://arxiv.org/abs/2312.11707}, 
}

@misc{khrulkov2020hyperbolic,
      title={Hyperbolic Image Embeddings}, 
      author={Valentin Khrulkov and Leyla Mirvakhabova and Evgeniya Ustinova and Ivan Oseledets and Victor Lempitsky},
      year={2020},
      eprint={1904.02239},
      archivePrefix={arXiv},
      primaryClass={cs.CV},
      url={https://arxiv.org/abs/1904.02239}, 
}

@article{kansal2023evaluating,
   title={Evaluating generative models in high energy physics},
   volume={107},
   ISSN={2470-0029},
   url={http://dx.doi.org/10.1103/PhysRevD.107.076017},
   DOI={10.1103/physrevd.107.076017},
   number={7},
   journal={Physical Review D},
   publisher={American Physical Society (APS)},
   author={Kansal, Raghav and Li, Anni and Duarte, Javier and Chernyavskaya, Nadezda and Pierini, Maurizio and Orzari, Breno and Tomei, Thiago},
   year={2023},
   month=Apr }

@misc{gemici2016normalizing,
      title={Normalizing Flows on {Riemannian} Manifolds}, 
      author={Mevlana C. Gemici and Danilo Rezende and Shakir Mohamed},
      year={2016},
      eprint={1611.02304},
      archivePrefix={arXiv},
      primaryClass={stat.ML},
      url={https://arxiv.org/abs/1611.02304}, 
}

@article{diepeveen2025curvaturecorrected,
    author = {Diepeveen, Willem and Chew, Joyce and Needell, Deanna},
    title = {Curvature-corrected tangent space-based approximation of manifold-valued data},
    journal = {Information and Inference: A Journal of the IMA},
    volume = {14},
    number = {4},
    pages = {iaaf031},
    year = {2025},
    month = {12},
    issn = {2049-8772},
    doi = {10.1093/imaiai/iaaf031},
    url = {https://doi.org/10.1093/imaiai/iaaf031},
    eprint = {https://academic.oup.com/imaiai/article-pdf/14/4/iaaf031/65761890/iaaf031.pdf},
}

@Inbook{Barth2025,
author="Barth, Lukas Silvester
and Fahimi, Hannaneh
and Joharinad, Parvaneh
and Jost, J{\"u}rgen
and Keck, Janis",
title="Metrics and {Riemannian} Manifolds",
bookTitle="Data Visualization with Category Theory and Geometry: With a Critical Analysis and Refinement of UMAP",
year="2025",
publisher="Springer Nature Switzerland",
address="Cham",
pages="25--104",
isbn="978-3-031-97973-6",
doi="10.1007/978-3-031-97973-6_3",
url="https://doi.org/10.1007/978-3-031-97973-6_3"
}

@article{Porcu02042016,
author = {Emilio Porcu and Moreno Bevilacqua and Marc G. Genton},
title = {Spatio-Temporal Covariance and Cross-Covariance Functions of the Great Circle Distance on a Sphere},
journal = {Journal of the American Statistical Association},
volume = {111},
number = {514},
pages = {888--898},
year = {2016},
publisher = {Taylor \& Francis},
doi = {10.1080/01621459.2015.1072541},


URL = { 
    
        https://doi.org/10.1080/01621459.2015.1072541
    
    

},
eprint = { 
    
        https://doi.org/10.1080/01621459.2015.1072541
    
    

}

}

@article{jeong2017spherical,
  author    = {Jaehong Jeong and Mikyoung Jun and Marc G. Genton},
  title     = {Spherical Process Models for Global Spatial Statistics},
  journal   = {Statistical Science},
  volume    = {32},
  number    = {4},
  pages     = {501--513},
  year      = {2017},
  publisher = {Institute of Mathematical Statistics},
  doi       = {10.1214/17-STS620},
  url       = {https://doi.org/10.1214/17-STS620}
}

@inproceedings{ramdas2015decreasing,
  title={On the decreasing power of kernel and distance based nonparametric hypothesis tests in high dimensions},
  author={Ramdas, Aaditya and Reddi, Sashank Jakkam and P{\'o}czos, Barnab{\'a}s and Singh, Aarti and Wasserman, Larry},
  booktitle={Proceedings of the AAAI Conference on Artificial Intelligence},
  volume={29},
  number={1},
  year={2015}
}

@misc{bu2026ggballgraphgenerativemodel,
      title={{GGBall}: Graph Generative Model on {Poincar\'e} Ball}, 
      author={Tianci Bu and Chuanrui Wang and Hao Ma and Haoren Zheng and Xin Lu and Tailin Wu},
      year={2026},
      eprint={2506.07198},
      archivePrefix={arXiv},
      primaryClass={cs.LG},
      url={https://arxiv.org/abs/2506.07198}, 
}

\clearpage
\appendix

\section{Related work}\label{ap:related_work}

Riemannian generative models build the geometry of the data space into the model, using densities, vector fields, diffusion processes, or probability paths on manifolds; examples include Riemannian normalizing flows \citep{gemici2016normalizing,mathieu2020riemannian}, Riemannian diffusion models \citep{de2022riemannian}, and Riemannian Flow Matching (RFM; \citet{chen2023flow}).
Such models are used for non-Euclidean data such as spherical Earth data \citep{mathieu2020riemannian}, hyperbolic image representations \citep{khrulkov2020hyperbolic}, rotations and structured manifolds \citep{jagvaral2024unified}, and molecular or protein-design tasks \citep{chen2023flow,zaghen2025riemannian}.

Our work instead studies how generated manifold samples should be evaluated.
Related work by \citet{kansal2023evaluating} evaluates generative modelling metrics in high-energy physics and show that different two-sample criteria vary in their sensitivity to controlled failure modes.
We extend this evaluation question to controlled constant-curvature manifolds, where curvature, dimension, volume scaling, target structure, and the choice of distance representation can all affect metric reliablity.

\textbf{Riemannian Flow Matching: extended.}\quad
Unlike diffusion models that often require many stochastic sampling steps, Flow Matching (FM) trains continuous normalizing flows (CNFs; \citet{chen2018}) by directly regressing a time-dependent vector field along conditional probability paths, avoiding simulation during training \citep{lipman2022flow}. 
RFM extends this framework to non-Euclidean geometries, making geometry an explicit inductive bias \citep{chen2023flow}.

RFM has been initially applied to data defined on spheres, tori, and mesh manifolds \citep{chen2023flow}, showing simulation-free training is not limited to Euclidean space.
As RFM computes its loss in tangent space, it is a local linear approximation that does not fully account for curved geometry.
Extensions such as Riemannian Gaussian Variational FM (RG-VFM; \citet{zaghen2025riemannian}) replace RFM's tangent-space loss with endpoint-based losses that minimise geodesic distances between predicted and target points.
By doing so, RG-VFM improves performance in material and protein design without modifying the underlying network architecture \citep{zaghen2025riemannian}.

In addition to the formal introduction of RFM in Section~\ref{sec:preliminaries}, RFM is trained through Riemannian Conditional Flow Matching (RCFM), which trains $v_t$ to match $\dot{\boldsymbol{x}}_t$ along the conditional path:
\begin{equation}\label{eq:RCFM_loss}
\mathcal{L}_{\mathrm{RCFM}}(\boldsymbol{\theta}) = \mathbb{E}_{t \sim \mathcal{U}(0,1),\, x_0 \sim p_0,\, x_1 \sim p_1}\!{\left[ \|\boldsymbol{v}_t(x_t; \boldsymbol{\theta}) - \dot{\boldsymbol{x}}_t\|^2_{g_\kappa} \right]}.
\end{equation}
\section{Geometric background}\label{ap:geometric_background}

We provide additional background on constant-curvature spaces $\mathcal{M}^d_\kappa$, where sectional curvature $\kappa$ determines how geometry differs from flat Euclidean space. In geodesic polar coordinates around a fixed origin $o$, all curvature effects are captured a single radial scaling term $\mathrm{sn}_\kappa(r)$ in Eq.~\eqref{eq:S_k}, enabling direct comparison across geometries $\kappa$~\citep{petersen2006riemannian}. As derived in Appendix~\ref{ap:volume_growth}, the scaling function $\mathrm{sn}_\kappa(r)$ governs volume growth. It is linear for $\kappa = 0$, bounded for $\kappa > 0$, and exponential for $\kappa < 0$. This is reflected in the Riemannian volume element $d\text{vol}_\kappa = \mathrm{sn}_\kappa(r)^{d-1} \, dr \, d\Omega$.

For hyperbolic space ($\kappa < 0$), we use the Poincaré ball model, which admits closed-form geodesics and exponential maps. In contrast, for $\kappa >0$, the exponential map at the origin is only locally diffeomorphic; beyond the injectivity radius $r = \pi/\sqrt{\kappa}$, geodesics converge and the spherical parameterisation is no longer globally one-to-one \citep{petersen2006riemannian}.

The exponential and logarithmic maps at $o$ can be further specified as \citep{chen2023flow}:
\begin{equation}\label{eq:exp_and_log_map}
    \exp_o^\kappa:T_o\mathcal{M}_\kappa^d\to\mathcal{M}_\kappa^d,
    \qquad
    \log_o^\kappa:\mathcal{M}_\kappa^d\to T_o\mathcal{M}_\kappa^d,
\end{equation}
The exponential map takes a tangent vector $\boldsymbol{v} \in T_o\mathcal{M}_\kappa^d$ and maps it to the manifold point reached by following the geodesic from $o$ in direction $\boldsymbol{v}$.
The logarithmic map does the opposite: for a point $x \in \mathcal{M}_\kappa^d$, it gives the tangent vector at $o$ that points from $o$ to $x$.
Thus (for points where this reverse map is well-defined), $x=\exp_o^\kappa(v) \Longleftrightarrow v=\log_o^\kappa(x)$.
In our experiments, we use $\exp_o^\kappa$ to map tangent-space samples to the manifold, and $\log_o^\kappa$ to map samples back to $T_o\mathcal{M}_\kappa^d$ for, e.g., evaluation on the tangent space, as a shared space for different curvatures.

\clearpage

\subsection{Tangent-space normalisation by curvature radius}\label{ap:tangent_space_normalisation}

In Section~\ref{sec:distributions}, we introduce tangent-space normalisation for distributions $\tilde p_0,\tilde p_1$ before their pushforward to the manifold through the exponential map.
Here, we give context on why this normalisation is appropriate for studying RFM behaviour beyond curvature-induced geometric scaling.

With constant-curvature manifolds, the curvature parameter induces, besides, e.g., volume scaling (Section~\ref{sec:geometries}, Appendix~\ref{ap:metrics_volume scaling}), a
natural length scale: a sphere of radius $R$ has constant sectional
curvature $\kappa = R^{-2}$ and a hyperboloid of radius $R$ has constant
sectional curvature $\kappa = -R^{-2}$ \citep{lee1997riemannian}.
Solving these relations for $R$ and by using $|\kappa|$ to combine both cases, we get curvature radius 
\begin{equation}\label{eq:curvature_radius}
    R_\kappa := |\kappa|^{-1/2}.
\end{equation}
(This same scale is also used in curvature-parameterised Poincar\'e (see, e.g., \citealp{ganea2018hyperbolic, van2023poincare}) and $\kappa$-stereographic models (see, e.g., \citet{bachmann2020constant}), where the radius is proportional to $|\kappa|^{-1/2}$.)
Thus, a fixed tangent-space radius $r$ does not have the same relative meaning across curvatures.
If $r$ is kept fixed while $|\kappa|$ increases, then $\frac{r}{R_\kappa} = \frac{r}{|\kappa|^{-1/2}} = \sqrt{|\kappa|} \, r$ also increases.
In other words, the same tangent displacement has the same absolute length $r$, but not the same relative length: when $|\kappa|$ increases, $R_\kappa$ becomes smaller, so the same $r$ ``reaches farther'' onto the manifold.
Therefore, RFM performance differences may not come from just curvature, but also from the distributions being placed at a different relative scale across different $\kappa$.

To isolate the effect of curvature more, we hence compare distributions at the same radius measured in units of the curvature radius (rather than at the same ``raw'' tangent-space radius).
We define the dimensionless radius
\begin{equation}\label{eq:dimensionless_radius}
    \rho
    :=
    \frac{r}{R_\kappa}
    =
    \sqrt{|\kappa|} \, r,
\end{equation}
which, by solving for $r$, gives the corresponding
\emph{curvature-dependent} radius
\begin{equation}\label{eq:curvature_scaled_radius}
    r_\kappa
    :=
    R_\kappa \rho
    =
    \frac{\rho}{\sqrt{|\kappa|}},
\end{equation}
so that when $|\kappa|$ increases, $r_\kappa$ is reduced, keeping $\rho$ constant.
Practically, keeping $\rho$ fixed means that the distribution is placed at the same relative distance from the origin across different $\kappa$.

In our set-up, this rescaling is applied to samples on the tangent space $z_i \sim \tilde p_i$, giving $\tilde{z}_\kappa = z/\sqrt{|\kappa|}$ if $\kappa\neq 0$, and $\tilde{z}_\kappa = z$ otherwise.
Thus, this operation rescales the data distribution on the tangent space relative to curvature radius (as opposed to scaling the Riemannian metric itself \citep{you2026constant}), and it does so \emph{before} applying the pushforward to the manifold.

We can see the effect of the normalisation by applying it to/through $\mathrm{sn}_\kappa$, defined as Eq.~\eqref{eq:S_k}.
For $\kappa\neq 0$, substituting Eq.~\eqref{eq:curvature_scaled_radius} into Eq.~\eqref{eq:S_k} gives
\begin{equation}
    \mathrm{sn}_\kappa(r_\kappa)
    =
    \begin{cases}
        \dfrac{\sin(\sqrt{\kappa}\,R_\kappa\rho)}{\sqrt{\kappa}}
        =
        \dfrac{\sin(\rho)}{\sqrt{\kappa}}
        =
        R_\kappa\sin(\rho)
        & \text{if } \kappa>0;\\[1.2em]
        \dfrac{\sinh(\sqrt{-\kappa}\,R_\kappa\rho)}{\sqrt{-\kappa}}
        =
        \dfrac{\sinh(\rho)}{\sqrt{-\kappa}}
        =
        R_\kappa\sinh(\rho)
        & \text{if } \kappa<0.
    \end{cases}
    \label{eq:S_k_scaled}
\end{equation}
Thus, after normalisation with the curvature radius, the dependence on $|\kappa|$ is isolated in $R_\kappa$.
By the same substitution, now into $J_\kappa$ (as defined in Eq.~\eqref{eq:Jacobian_factor}) using Eq.~\eqref{eq:curvature_scaled_radius} and Eq.~\eqref{eq:S_k_scaled}, we get
\begin{equation}
    J_\kappa(r_\kappa)
    =
    \left(
        \frac{\mathrm{sn}_\kappa(r_\kappa)}{r_\kappa}
    \right)^{d-1}
    =
    \begin{cases}
        \left(
            \dfrac{R_\kappa\sin(\rho)}
                  {R_\kappa\rho}
        \right)^{d-1}
        =
        \left(
            \dfrac{\sin(\rho)}{\rho}
        \right)^{d-1}
        & \text{if } \kappa>0;\\[1.2em]
        \left(
            \dfrac{R_\kappa\sinh(\rho)}
                  {R_\kappa\rho}
        \right)^{d-1}
        =
        \left(
            \dfrac{\sinh(\rho)}{\rho}
        \right)^{d-1}
        & \text{if } \kappa<0.
    \end{cases}
    \label{eq:Jacobian_dimensionless}
\end{equation}
(For $\kappa=0$, no normalisation is applied, giving $J_0(r)=1$.)
As $R_\kappa$ cancels for $\kappa\neq 0$, the direct dependence on curvature magnitude is removed.
Nonetheless, the volume behaviour dependent on the sign remains (through $\sin$ and $\sinh$, resp.), meaning the set-up controls for global, curvature-induced scaling, but does not fully normalise/control for curvature effects.

\subsection{Volume growth in constant curvature}\label{ap:volume_growth}

This Appendix supports the volume scaling diagnostic used in Section~\ref{sec:geometries} and Appendix~\ref{ap:metrics_volume scaling}.
We show how constant curvature changes the volume available at geodesic radius $r$: Euclidean volume grows as $r^d$, hyperbolic volume grows exponentially, and spherical volume is bounded.
This explains why the same target distribution can experience different curvature-induced volume distortion with different geometries and dimensions.

\textbf{Hyperbolic space $\mathbb{H}^n(-c)$.}\quad For 2D surface areas, the circumference at a geodesic distance $\rho$ in a space of constant negative curvature $-c$ is:
\begin{equation}
    \mathbb{C}(\rho) = 2\pi \frac{\sinh(\sqrt{c}\,\rho)}{\sqrt{c}}.
\end{equation}
The total area is the accumulation of these circumferences from $0$ to $r$:
\begin{equation}
    A_{\mathbb{H}}(r) = \int_{0}^{r} 2\pi \frac{\sinh(\sqrt{c}\,\rho)}{\sqrt{c}} \, d\rho = \frac{2\pi}{\sqrt{c}} \int_{0}^{r} \sinh(\sqrt{c}\,\rho) \, d\rho.
\end{equation}
Using the identity $\int \sinh(ax) \, dx = \frac{1}{a} \cosh(ax)$, we find
\begin{align}
    A_{\mathbb{H}}(r) &= \frac{2\pi}{c} \left( \cosh(\sqrt{c}\,r) - 1 \right),\\
    A_{\mathbb{H}}(r) &= \frac{2\pi}{c} \left( \frac{e^{\sqrt{c}\,r} + e^{-\sqrt{c}\,r}}{2} - 1 \right).
\end{align}

As the radius increases ($r \rightarrow \infty$), the fundamental properties of the manifold's geometry become dominant. For larger $r$, the $-1$ term becomes negligible and $e^{-\sqrt{c}\,r}$ approaches $0$. Thus, we find that the area in hyperbolic space
\begin{equation}
    A_{\mathbb{H}}(r) \approx \frac{\pi}{c} e^{\sqrt{c}\,r},
\end{equation}
grows exponentially relative to the radius, with the curvature constant $c$ acting as a scaling factor that accelerates the expansion of available space.

For higher dimensional spaces ($n > 2$), the surface area of an $(n-1)$-sphere of geodesic radius $\rho$ is given by:
\begin{equation}
    S_{n-1}(\rho) = \Omega_{n-1} \left( \frac{\sinh(\sqrt{c}\,\rho)}{\sqrt{c}} \right)^{n-1},
\end{equation}
where $\Omega_{n-1}$ denotes the surface area of the unit $(n-1)$-sphere.
The total volume $V_{\mathbb{H}}(r)$ is the integral of these ``shells'' from the center ($0$) to the total geodesic radius ($r$) is
\begin{equation}
    V_{\mathbb{H}^n}(r)
    = \int_{0}^{r} S_{n-1}(\rho)\,d\rho
    = \frac{\Omega_{n-1}}{(\sqrt{c})^{n-1}}
    \int_{0}^{r} \sinh^{n-1}(\sqrt{c}\,\rho)\,d\rho.
\end{equation}

Thus, in hyperbolic space, as shown in the 2D example, the $\sinh$-function ensures that the surface area of each subsequent shell increases, leading to the exponential volume growth characteristic of negative curvature
\begin{equation}
    V_{\mathbb{H}^n}(r) \propto e^{(n-1)\sqrt{c}r} \quad (r\to\infty).
\end{equation}

\textbf{Spherical space $\mathbb{S}^2(+c)$.}\quad In a space of constant positive curvature $+c$, the geometry ``closes'' in on itself. For 2D surface areas, the circumference at a geodesic distance $\rho$ is
\begin{equation}
    \mathbb{C}(\rho) = 2\pi \frac{\sin(\sqrt{c}\,\rho)}{\sqrt{c}}.
\end{equation}
The total area $A_{\mathbb{S}}$ is the accumulation of these circumferences:
\begin{equation}
    A_{\mathbb{S}}(r) = \int_{0}^{r} 2\pi \frac{\sin(\sqrt{c}\,\rho)}{\sqrt{c}} \, d\rho = \frac{2\pi}{c} \left( 1 - \cos(\sqrt{c}\,r) \right).
\end{equation}

Given $-1 \leq \cos(x) \leq 1$, we find
\begin{align}
    0 \leq 1-\cos(\sqrt{c}\, r) \leq 2,\\
    0 \leq \frac{2\pi}{c} \left( 1-\cos(\sqrt{c}\ r) \right) \leq \frac{4\pi}{c}, \\
    0 \leq A_{\mathbb{S}}(r) \leq \frac{4\pi}{c}.
\end{align}
Consequently, in spherical space, the area growth is strictly bounded by the manifold's closed geometry. We find $A_{\mathbb{S}, \text{max}} = \frac{4\pi}{c}$  for $r=\frac{\pi}{\sqrt{c}}$ (antipode), which is the area of the full sphere of radius $1/\sqrt{c}$.

For higher-dimensional spaces ($n>2$) and $r\leq \pi/\sqrt{c}$, the surface area of an ($n-1$)-sphere of geodesic radius $\rho$ is given by
\begin{equation}
    S_{n-1}(\rho) = \Omega_{n-1} \left( \frac{\sin(\sqrt{c}\,\rho)}{\sqrt{c}} \right)^{n-1}.
\end{equation}
The total volume $V_{\mathbb{S}^n}(r)$ is calculated by integrating these spherical shells:
\begin{align}
    V_{\mathbb{S}^n}(r)
    &= \frac{\Omega_{n-1}}{c^{n/2}} \int_{0}^{\sqrt{c}\,r} \sin^{n-1}(t)\,dt \\
    &< \frac{\Omega_{n-1}}{c^{n/2}} \int_{0}^{\sqrt{c}\,r} t^{n-1}\,dt
     = \frac{\Omega_{n-1}}{c^{n/2}} \cdot \frac{(\sqrt{c}\,r)^n}{n}
     = \frac{\Omega_{n-1}\, r^n}{n}.
\end{align}

Because $\sin t < t$ on $(0, \pi]$, we get $(V_{\mathbb{S}^n}(r) < \Omega_{n-1}r^n/n$, the Euclidean volume, so spherical volume growth is sub-Euclidean.
At $r=\pi/\sqrt{c}$ the integral gives $V = \Omega_nc^{-n/2}$, the volume of the whole sphere.
\begin{figure}[b]
    \centering
    \includegraphics[width=1\linewidth]{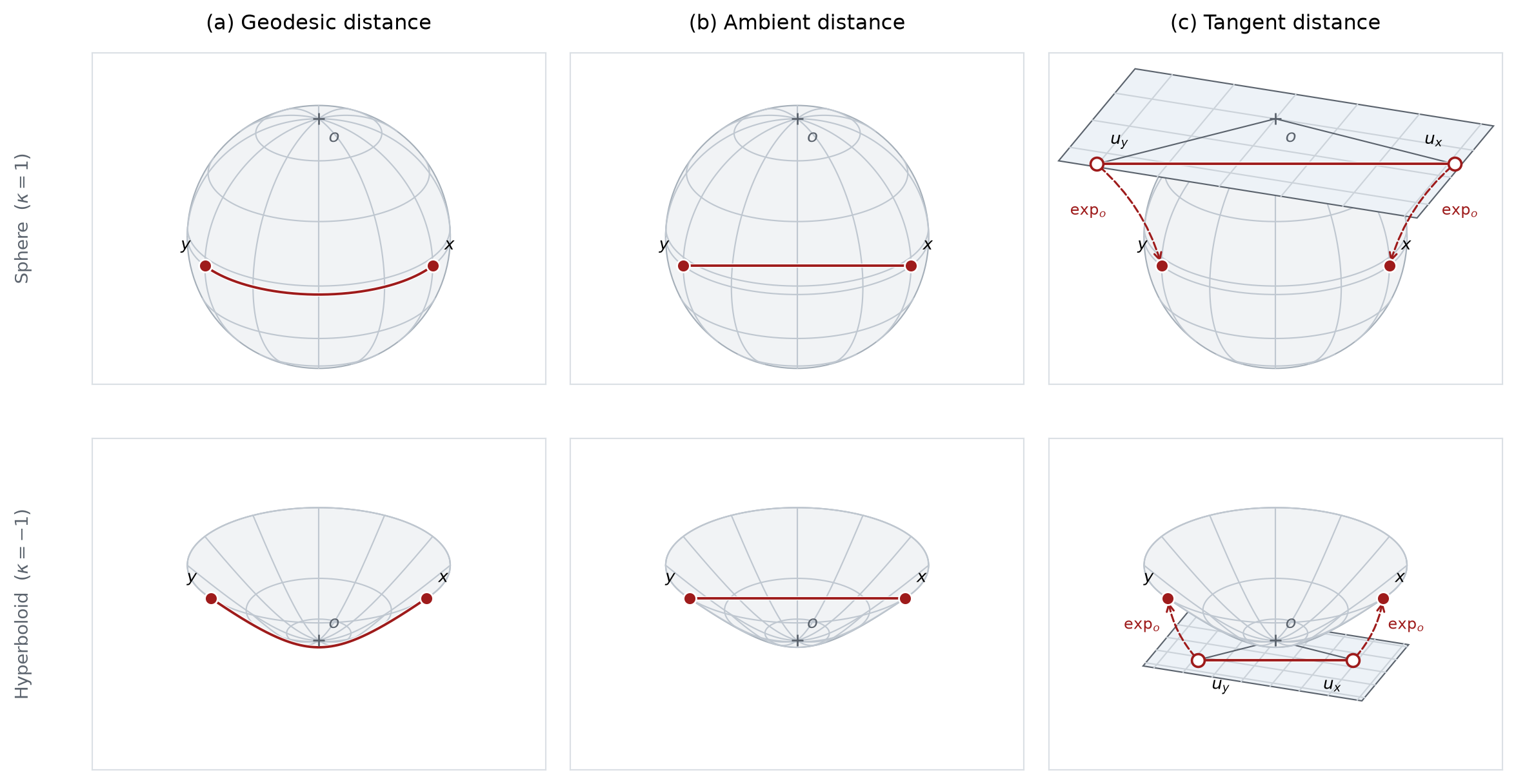}
    \caption{The three evaluation distances, illustrated for spherical and hyperbolic space. Samples are compared intrinsically by geodesic distance, extrinsically by ambient Euclidean distance, or locally after log-mapping to the tangent space at $o$. The same pair of points can thus produce different error.}
    \label{fig:evaluation_distances}
\end{figure}

\section{Evaluation metrics}\label{ap:metrics}

Here we provide additional intuition, motivation, and details for the metrics introduced in Section~\ref{sec:metrics}.
We also introduce alternative metrics that were not included in the main text due to space constraints.
We divide these metrics into categories of accuracy (\ref{ap:metrics_accuracy}), stability (\ref{ap:metrics_stability}), and volume scaling (\ref{ap:metrics_volume scaling}).

On manifolds, the evaluation distance is a modelling choice.
Embedded samples can be compared intrinsically by geodesic distance or extrinsically by ambient Euclidean distance, and these generally differ in a curvature-dependent way \citep{Barth2025}; similar distance choices are standard in spherical spatial statistics, see e.g.\ \citet{Porcu02042016, jeong2017spherical}.
Likewise, tangent-space distances provide a useful Euclidean approximation after log mapping, but such local comparisons can miss global curvature effects \citep{diepeveen2025curvaturecorrected}.
We therefore evaluate $d_{\mathrm{geodesic}}$ (i.e., $d_\kappa$), $d_{\mathrm{ambient}}$, and $d_{\mathrm{tangent}}$ separately, illustrated by Figure~\ref{fig:evaluation_distances}.

\subsection{Accuracy}\label{ap:metrics_accuracy}

\begin{figure}[h]
    \centering
    \includegraphics[width=\linewidth]{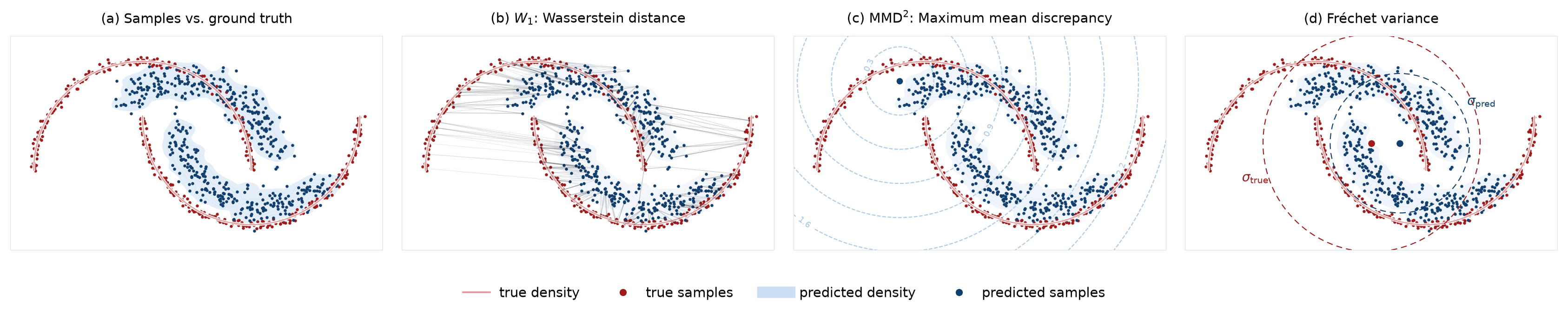}
    \caption{Illustration of the evaluation metrics. The same predicted samples can be assessed by transport cost ($W_1$), kernel similarity ($\mathrm{MMD}^2$), and dispersion via Fr\'echet variance.}
    \label{fig:metric_illustration}
\end{figure}

Figure~\ref{fig:metric_illustration} illustrates the complementary roles of our three evaluation metrics.

\paragraph{Wasserstein distance.}
We measure how well $\hat{p}_1$ matches $p_1$ using the $1$-Wasserstein distance $W_1(p_1,\hat{p}_1)$ on $\mathcal{M}_\kappa^d$ defined in Eq.~\eqref{eq:Wasserstein} in Section~\ref{sec:metrics}.
This distance is sensitive to the underlying metric structure of the space.
On curved manifolds, this makes Wasserstein distances a more natural choice than KL-divergence (relative entropy), which is based on pointwise density ratios.
In our experiments, we approximate $W_1$ with entropy-regularized optimal transport and use the debiased Sinkhorn divergence
\citep{genevay2018learning,feydy2019interpolating}
\begin{equation}
S_\varepsilon(p_1,\hat{p}_1)
=
\mathrm{OT}_\varepsilon(p_1,\hat{p}_1)
- \tfrac{1}{2}\,\mathrm{OT}_\varepsilon(p_1,p_1)
- \tfrac{1}{2}\,\mathrm{OT}_\varepsilon(\hat{p}_1,\hat{p}_1),
\end{equation}
where $\mathrm{OT}_\varepsilon$ denotes the entropy-regularized  
OT cost.
This regularization makes the transport problem smoother and easier to solve numerically, but it also introduces bias.
The two correction terms remove this bias while keeping the computational advantages.

\textbf{Dispersion.}\quad 
To detect over- or underdispersion of $p_1$ and $\hat{p}_1$ on $\mathcal{M}_\kappa^d$, we use the Fr\'echet variance
\begin{equation}\label{eq:Frechet_variance}
    \sigma^2_F(p) := \mathbb{E}_{x\sim p}\!\left[d_\kappa(\mu_F, x)^2\right], \qquad \mu_F(p) := \operatorname{arg\min}_{m \in \mathcal M_\kappa^d}\, \mathbb{E}_{x\sim p} \! \left[d_\kappa(m, x)^2\right],
\end{equation}
where $\mu_F$ is the Fr\'echet mean (also known as Riemannian centre of mass) \citep{afsari2011riemannian, karcher2014riemannian}.
We compare target and generated $\sigma^2_F$ using their ratio $\rho_F(\hat{p}_1, p_1):= \sigma^2_F(\hat{p}_1)/\sigma^2_F(p_1)$.

\textbf{Specifically for MoG experiment: mode coverage.}\quad
We additionally measure whether generated samples cover the target modes in the correct proportions for the MoG experiments.
Each generated sample is assigned to its nearest target mode center using the intrinsic distance $d_\kappa$, giving an empirical mode distribution $\hat p$, whose $k$-th entry is the fraction of generated samples assigned to mode $k$.
We compare this to the target mixture weights $p$ using
\begin{equation}\label{eq:KL_divergence_MoG}
    D_{\mathrm{KL}}(p \,\|\, \hat p)
    =
    \sum_{k=1}^K p_k \log \frac{p_k}{\hat p_k+\epsilon}.
\end{equation}


\subsection{Stability}\label{ap:metrics_stability}

We evaluate stability using three metrics:
\begin{itemize}[leftmargin = 20pt, topsep = -1pt, itemsep = -1pt]
    \item[(\emph{i})] We track vector-field magnitudes through the mean, standard deviation, maximum, and 95th percentile of $\|v_t(x)\|_{g_\kappa}$.
    Large or heavy-tailed norms indicate that the learned flow requires strong local velocities, which can make both training and sampling less stable numerically.
    \item[(\emph{ii})] Since $v_t(x)$ should lie in $T_x\mathcal{M}_\kappa^d$, we measure tangent-space violation by projecting $v_t(x)$ back onto the tangent space and recording the remaining non-tangent component.
    We do this both for its absolute size and relative to $\|v_t(x)\|_{g_\kappa}$.
    For the sphere, this corresponds to checking whether $v_t(x)$ is orthogonal to $x$, i.e.\ whether $\langle x,v_t(x)\rangle=0$.
    \item[(\emph{iii})] Finally, we report the fraction of finite samples, i.e.\ the proportion of samples whose coordinates contain no \texttt{NaN}s or infinities.
    \end{itemize}

\begin{wrapfigure}[10]{r}{0.4\textwidth}
    \vspace{-2.5em}
    \centering
    \includegraphics[width=\linewidth]{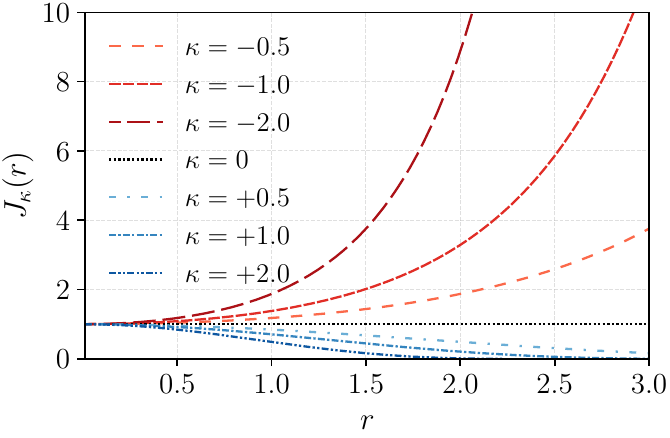}
    \vspace{-2em}
    \caption{$J_\kappa$ versus $r$ for different $\kappa$.}
    \label{fig:theoretical_Jacobian_factor}
    \vspace{-1.2em}
\end{wrapfigure}
\subsection{Volume scaling.}\label{ap:metrics_volume scaling}
To quantify volume scaling for a distribution $p$ on $\mathcal{M}_\kappa^d$, we use the radial Jacobian of the exponential map; $J_\kappa(r) := \left|\det D_{\boldsymbol{v}} \exp_o^\kappa\right|$.
In geodesic polar coordinates around a fixed origin $o$, the Riemannian volume element follows $d\text{vol}_\kappa = \mathrm{sn}_\kappa(r)^{d-1} \, dr \, d\Omega$,   
in which $dr$ and $d\Omega$ denote infinitesimal radial and angular elements, respectively. 
In the flat tangent space $T_o\mathcal{M}_\kappa^d$ (viewed as $\mathbb{R}^d$), the corresponding Euclidean volume element is $r^{d-1}\,dr\,d\Omega$ \citep{gallot2004riemannian}.
Their ratio gives the Jacobian factor (Eq.~\eqref{eq:Jacobian_factor}), which measures the local volume ``distortion'' induced by the geometry('s curvature) at geodesic radius $r$ \citep{lang2020riemannian}, see Figure~\ref{fig:theoretical_Jacobian_factor}.

Given a distribution $p$ on $\mathcal{M}_\kappa^d$, we summarise how much curvature-induced volume scaling it experiences by averaging this factor over points $x$ drawn from $p$, $x \sim p$, as in Eq.~\eqref{eq:mean_volume_growth}.
In other words, this turns a quantity for a point into a summary for a distribution, by taking the expectation of the quantity over the distribution.
If $\mu_{J_\kappa}(p)$ is large, then the probability mass of $p$ lies (on average) where the geometry expands volume; if it is small, then the mass lies where the geometry contracts volume.
Hence, it summarises how much volume growth the distribution ``experiences.''
Additionally, we also consider the variance $\sigma^2_{J_\kappa}(p)$ of the Jacobian, i.e.\
\begin{equation}\label{eq:Jacobian_variance}
\sigma^2_{J_\kappa}(p)
:=
\operatorname{Var}_{x\sim p}\!\left[J_\kappa(r(x))\right].
\end{equation}

We further define, given a start distribution $p_0$ and a target distribution $p_1$, the \emph{volume-growth gap} as 
\begin{equation}\label{eq:volume_growth_gap}
    \Delta \mu_{ J_\kappa}(p_0, p_1) := \mu_{ J_\kappa}(p_1) - \mu_{ J_\kappa}(p_0),
\end{equation}
which, intuitively, measures how much volume growth the flow must overcome between $p_0$ and $p_1$.
\begin{figure}
    \centering
    \includegraphics[width=0.85\linewidth]{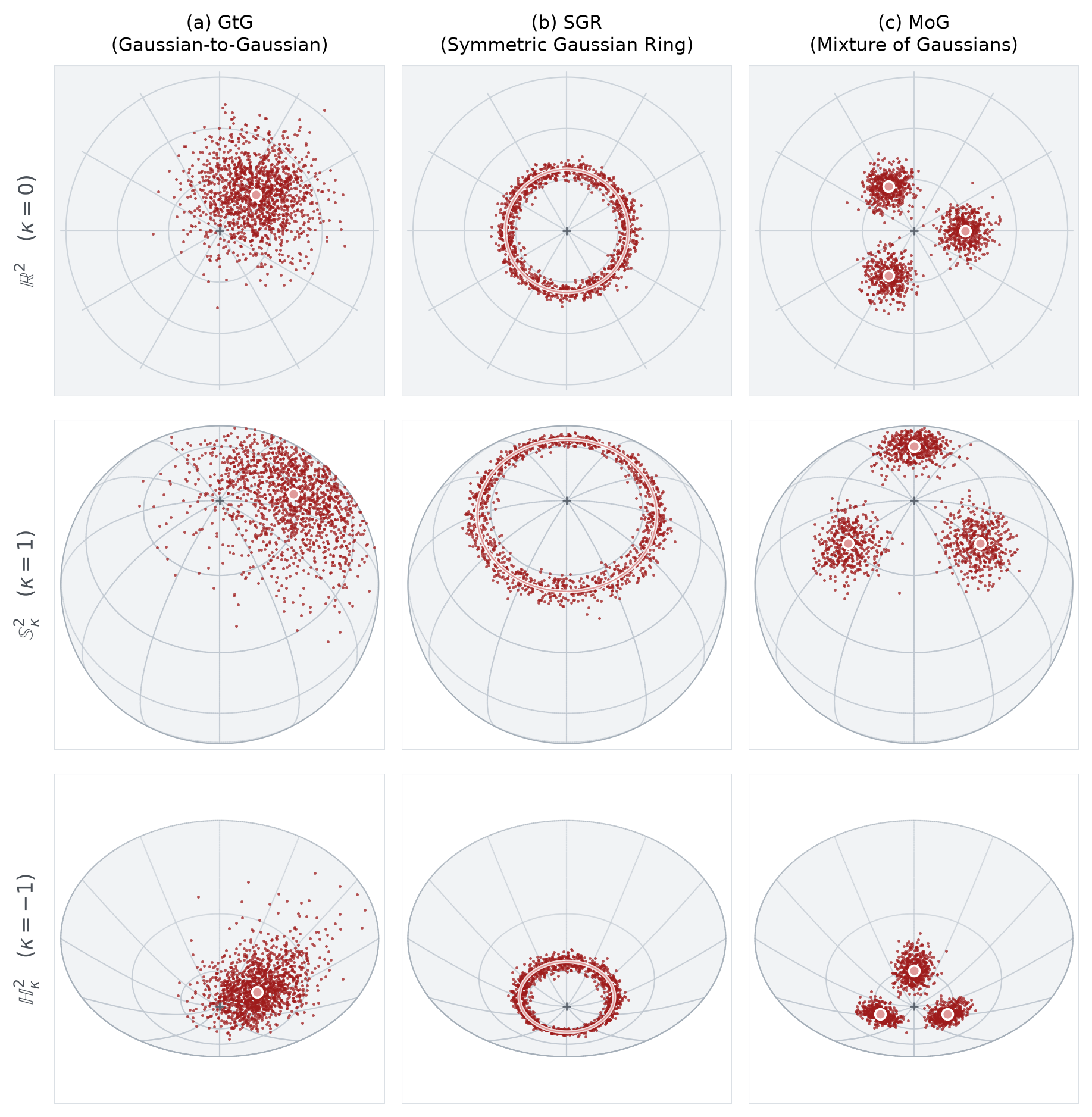}
    \caption{Overview of the target distributions. Columns show GtG, SGR, and MoG targets; rows show Euclidean, spherical, and hyperbolic geometries. The same tangent-space distributions lead to different sample geometries after mapping to manifolds with different curvature.}
    \label{fig:experimental_families}
\end{figure}

\section{Experimental setup}\label{ap:experimental_setup}
This Appendix gives more detailed explanation on the experimental setup and hyperparameters for both the RFM experiments, with the Gaussian-to-Gaussian (\ref{ap:GtG}), Gaussian-ring (\ref{ap:Gaussian_ring}), and Mixture-of-Gaussians (\ref{ap:MoG}), and also the noise-perturbation tests (\ref{ap:noise}).
Figure~\ref{fig:experimental_families} visualises the three target families.

\begin{table}[h]
\vspace{-1em}
\centering
\caption{General hyperparameters.}
\label{tab:general_hyperparameters}
\footnotesize
\begin{tabular}{ll}
\toprule
\textbf{Parameter} & \textbf{Value} \\
\midrule
Curvatures ($\kappa$) & -10, -5, -3, -2, -1, 0, 1, 2, 3, 5, 10  \\
Dimensions ($d$) & 2, 3, 4, 8, 16, 32, 64, 256, 512  \\
Seeds & $34,\ 42,\ 50$ \\
\bottomrule
\end{tabular}
\vspace{-1em}
\end{table}

\subsection{Gaussian-to-Gaussian}\label{ap:GtG}

\textbf{Motivation and geometric intuition.}\quad
To provide a baseline for the more complex distributions, we construct a Gaussian-to-Gaussian (GtG) matching task.
Both the base and target distributions are defined through isotropic Gaussian sampling in the tangent space of the manifold, followed by mapping to the manifold via the exponential map.

\textbf{Analytical formulation.}\quad
A tangent vector is sampled as $\boldsymbol{z} \sim \mathcal{N}(\boldsymbol{\mu}_o, \sigma_o^2)$ where $\boldsymbol{\mu} \in T_o\mathcal{M}_\kappa^d$ is the tangent-space mean and $\sigma$ is the isotropic standard deviation.
The sampled tangent vector is then mapped onto the manifold through the exponential map.
For the Poincar\'e ball model, the tangent vector is first projected to the tangent space and rescaled using the conformal factor at the origin before applying the exponential map.
When tangent normalisation is enabled (Appendix~\ref{ap:tangent_space_normalisation}), tangent vectors are additionally scaled by the curvature radius before being mapped to the manifold. 

\textbf{Mean placement strategy.}\quad
The source distribution is centered at the origin of the tangent space.
The target mean is constructed as a tangent vector with norm equal to the prescribed separation distance $\mu_1=\delta u$, where $\delta$ denotes the desired separation distance and $u$ is a unit vector in the tangent space.
For spherical manifolds, the tangent space at the north pole is orthogonal to the radial direction.
Therefore, the first coordinate of the tangent vector is constrained to zero.

\textbf{Varying complexity.}\quad
In addition to the general experimantal hyperparameters (see Table~\ref{tab:general_hyperparameters}), the hyperparameters as described in Table~\ref{tab:gtg_hyperparameters} are used. The separation distance or standard deviations are not varied across configurations.

\begin{table}[h]
\vspace{-1em}
\centering
\caption{Additional hyperparameters for the GtG experiments.}
\label{tab:gtg_hyperparameters}
\footnotesize
\begin{tabular}{ll}
\toprule
\textbf{Parameter} & \textbf{Value} \\
\midrule
Base standard deviation ($\sigma_0$) & 0.3 \\
Target standard deviation ($\sigma_1$) & 0.3 \\
Separation distance ($\delta$) & 0.5  \\
\bottomrule
\end{tabular}
\vspace{-1em}
\end{table}

\subsection{Symmetric Gaussian Ring (SGR)}\label{ap:Gaussian_ring}

\textbf{Motivation and geometric intuition.}\quad
The symmetric Gaussian ring (SGR) aims to construct rotationally symmetric experiments, such that we can isolate the effect of radial volume growth.
The ring places probability mass near radius $r_{\mathrm{SGR}}$ in all angular directions.
Equivalently, its radial coordinate is approx.\ Gaussian while its angular direction is uniform.
This results in a rotationally symmetric distribution that visually forms a ring shape for $d = 2$ and a hollow shell for $d > 2$.

\textbf{Analytical formulation}.\quad
To sample a point $\boldsymbol{z} \in T_o\mathcal{M}^d_\kappa$, we sample a radius $r \sim \mathcal{N}(r_\mathrm{SGR},\sigma_\mathrm{SGR}^2)$ and a unit direction $\hat{\boldsymbol{\theta}} \sim \mathcal{U}(\mathbb{S}^{d-1})$ in $T_o\mathcal M_\kappa^d$.
We then define the sample on the tangent space as $\boldsymbol{z}=r\hat{\boldsymbol{\theta}}$ and map it to the manifold by $x=\exp_o^\kappa(\boldsymbol{z})$.
This yields a rotationally symmetric target around the origin $o$, with probability mass concentrated near geodesic radius $r_\mathrm{SGR}$.
Varying $r_\mathrm{SGR}$ therefore changes the radial volume-growth ``experienced'' by the flow, as measured by Eq.~\eqref{eq:volume_growth_gap}.

\textbf{Varying complexity.}\quad
In addition to the general experimental hyperparameters of Table~\ref{tab:general_hyperparameters} and the $\sigma_0$ hyperparameter of Table~\ref{tab:gtg_hyperparameters}, we add the $r_\mathrm{SGR}$ and $\sigma_\mathrm{SGR}$ hyperparameters, as listed in Table~\ref{tab:gr_hyperparameters}.

For spherical geometries, radii must remain below the injectivity radius of the exponential map at $o$. With curvature-radius normalisation, this limit is $\rho<\pi$ for every $\kappa>0$. Our largest radius, $r_{\text{SGR}}=0.85$, corresponds to $0.85/\pi\approx 27\%$ of the injectivity radius, and with $\sigma_{\text{SGR}}=0.05$ essentially all target mass lies well inside it. For comparability, we use the same radii for Euclidean and hyperbolic geometries.

Here, complexity is varied by $r_\mathrm{SGR}$: with larger radii, the volume growth gap in Eq.~\eqref{eq:volume_growth_gap} increases, which is hypothesised to influence the complexity of the transport task.

\begin{table}[h]
\vspace{-1em}
\centering
\caption{Additional hyperparameters for the SGR experiments.}
\label{tab:gr_hyperparameters}
\footnotesize
\begin{tabular}{ll}
\toprule
\textbf{Parameter} & \textbf{Value} \\
\midrule
Radius ($r_\mathrm{SGR}$) & \num{0.35}, \num{0.60}, \num{0.85} \\
Radial standard deviation ($\sigma_\mathrm{SGR}$) & 0.05 \\
\bottomrule
\end{tabular}
\vspace{-1em}
\end{table}

\subsection{Mixture-of-Gaussians (MoG)}\label{ap:MoG}

\textbf{Motivation and geometric intuition.}\quad
MoG extends the GtG task to a multimodal setting.
The modes are placed symmetrically at equal geodesic distance from the source (analogous to the SGR's radial shell structure, but with discrete angular orientation instead of full rotational symmetry).
This breaks the isotropy that makes SGR ``relatively easy'' for RFM, requiring the flow to resolve distinct angular targets.
In principle, MoG approaches SGR in structure with sufficient modes, making it a natural interpolation between the two.
Complexity is controlled by the number of modes and the mixture weights, so we can isolate the effect of multimodality from that of radial volume growth.

\textbf{Analytical formulation.}\quad
We define the base distribution as an isotropic wrapped Gaussian $p_0 = \mathcal{N}_{\mathcal{M}}(\mu_0, \sigma_0^2)$ centered at the source mean $\mu_0$ on the manifold $\mathcal{M}$.
The target distribution is a mixture of $n$ Gaussians:
$$p_1 = \sum_{i=1}^n w_i \mathcal{N}_{\mathcal{M}}(\mu_i, \sigma_1^2),$$
where $\sum_{i=1}^n w_i = 1$.
To maintain parity with our GtG experiments and isolate the effect of constant sectional curvature, we strictly enforce that all target means $\mu_i$ are equidistant from $\mu_0$.
Given a reference target mean $\mu_{\mathrm{ref}}$ from the corresponding GtG setup, we constrain the geodesic distance such that $d_{\mathcal{M}}(\mu_0, \mu_i) = d_{\mathcal{M}}(\mu_0, \mu_{\mathrm{ref}})$ for all $i \in \{1, \dots, n\}$.

\textbf{Mean placement strategy.}\quad
To eliminate spatial bias, we set the target means to be maximally separated while satisfying the equidistance constraint.
We achieve this by formulating a generalised Thomson problem \citep{batle2014thomsonproblem} (i.e., we seek a set of $n$ vectors that minimise the inverse-distance electrostatic potential).

\textbf{Varying complexity.}
We ablate topological complexity along two axes (configuration details in Table~\ref{tab:mog_hyperparameters}):
\begin{itemize}[leftmargin = 20pt, topsep = -1pt, itemsep = -1pt]
    \item[(\emph{i})] \textbf{Mode scaling ($n \in \{2, 3, 5\}$):} Increases spatial crowding, requiring the model to resolve higher-order symmetric structures (e.g., regular simplices) in higher dimensions.
    \item[(\emph{ii})] \textbf{Weight distribution:} We compare balanced mixtures against unbalanced mixtures to test if the learned vector field can allocate proportional mass without mode collapse.
\end{itemize}

\begin{table}[h]
\vspace{-1em}
\centering
\caption{Additional hyperparameters for the MoG experiments.}
\label{tab:mog_hyperparameters}
\footnotesize
\begin{tabular}{ll}
\toprule
\textbf{Parameter} & \textbf{Value} \\
\midrule
Number of target modes ($n$) & 2, 3, 5 \\
Base standard deviation ($\sigma_0$) & 0.3 \\
Target standard deviation ($\sigma_1$) & 0.3 \\
Balanced weights ($w_n$) & $[1/n, \dots, 1/n]$ \\
Unbalanced weights ($w_n$) & $w_2=\frac{1}{3}[1,2], w_3=\frac{1}{4}[1,1,2], w_5=\frac{1}{8}[1,2,3,1,1]$ \\
\bottomrule
\end{tabular}
\end{table}

\textbf{Spherical injectivity exclusion for GtG and MoG.}\quad
For spherical GtG and MoG runs, the tangent Gaussian construction is only one-to-one inside the injectivity radius of the exponential map.
With our curvature normalisation this requires $\|z\|<\pi$.
Since the Gaussian standard deviation is $0.3$ per coordinate, this condition is violated in high dimensions: the probability of $\|z\|>\pi$ is below $0.1\%$ up to $d=64$, but essentially one for $d=256$ and $d=512$.
The samples are still valid points on the sphere, but no longer represent the intended one-to-one tangent Gaussian target.
We thus omit spherical GtG/MoG entries at $d\in\{256,512\}$ from all figures.
For instance, from the heatmaps of Figure~\ref{fig:geometric_metric_ratios_gtg}, \ref{fig:intrinsic_vs_nonintrinsic_curves}, \ref{fig:global_snr_heatmap}, \ref{fig:rq2_fr}, where hatched (grey with black diagonal lines) cells mark these omitted configurations.

\subsection{Noise experiment details}\label{ap:noise}
For each target configuration, we use $N=256$ samples and evaluate the noise levels
\begin{equation}\label{eq:noise_levels}
    \Sigma=\{0.005,0.01,0.02,0.05,0.10,0.20,0.40\}.
\end{equation}
The replicate index $s$ refers only to independent random seeds. Configurations with different mixture weights, numbers of mixture components, or ring radii are treated separately and are not pooled as replicates. Consecutive metric differences are paired by seed and target configuration.
Because we use three seeds, these SNR values should be interpreted as empirical reliability diagnostics rather than precise uncertainty estimates.
\section{Implementation details}\label{ap:implementation_details}

This section includes the practical implementation details used throughout the experiments, including architecture design, optimisation, numerical integration, deterministic evaluation, and geometry-specific stability design.
Our publicly available codebase is available on GitHub.\footnote{\url{https://github.com/MishaOberski/eval-riemannian-gen-modelling}}

\textbf{Software framework.}\quad
Our implementation extends the original RFM framework by \citet{chen2023flow} with additional components for deterministic evaluation, sampling in a curvature normalised way, and geometry-consistent evaluation.
We use code by \citet{valeau2026hyperbolic} for hyperbolic flow matching extensions.

All experiments are implemented using PyTorch and PyTorch lightning modules.
Manifold operations are based upon the \texttt{geoopt} module for Riemannian optimisation in PyTorch \citep{geoopt2020kochurov}.
We extend on this for curvature-aware dataset generation and evaluation.
All methods are executed on NVIDIA H100 Tensor Core GPUs using CUDA acceleration.\footnote{In total, we used an equivalent of approximately \num{490.5} H100 GPU hours.
Assuming an NVIDIA H100 power draw of \SI{700}{\watt} and using carbon intensity of $388\,\mathrm{g\,CO_2eq\,kWh^{-1}}$ \citep{nowtricity2026co2}, this gives
$133.22\,\mathrm{kgCO}_{2}\mathrm{eq}$.
This estimate excludes CPU, memory, storage, and (data-centre) overhead.}
Mixed-precision training is used where data is numerically stable. 
All experiments (i.e., training one flow matching model) took approx.\ 5 to 10 minutes using the above compute.

\textbf{Vector field architecture.}\quad
The learned transport field $v_\theta(t,x)$ is parameterised using a MLP operating in the ambient space of the given manifold.
The network takes the ambient coordinate representation of $x_t$ with a scalar time variable $t \in [0,1]$ as the input, where time conditioning is done by concatenation of $t$ to the input features.
The architecture consists of fully connected linear layers with SiLU activation, residual hidden blocks and a final projection layer. 
The output of the model is an unrestricted ambient vector field, which is projected onto the tangent bundle after every forward pass
\begin{equation}
    \boldsymbol{v}_\theta(t, x_t) = \operatorname{Proj}_{T_{x_t}\mathcal{M}_\kappa^d}\!\bigl(\tilde{\boldsymbol{v}}_\theta(t, x_t)\bigr),
\end{equation}
where $\tilde{\boldsymbol{v}}_\theta(t, x_t)$ denotes the raw ambient output of the network.
Importantly, we keep the hidden dimensionality and network depth fixed across geometries and experiments to isolate geometric effects from architectural scaling effects.

\textbf{Geodesic interpolation and velocity computation.}\quad
We compute conditional interpolation trajectories using the closed form exponential and logarithmic maps described in Eq.\ \eqref{eq:exp_log_maps}.

Instead of approximating tangent velocities using finite differences, we compute them using forward-mode automatic differentiation via Jacobian-vector-products (JVPs), this helps improve numerical stability on curved manifolds. 
To further stabilise the logarithmic map computation near the injectivity radius on the sphere manifold we utilise cosine clipping and small angle regularization. 
For the hyperbolic manifolds (i.e., the Poincar\'e ball model $\mathbb{B}^d_\kappa$, see e.g.\ \citet{valeau2026hyperbolic}), we use radius clipping near the boundary to avoid numerical overflow.
Lastly, we re-project tangent vectors to their respective tangent space to mitigate accumulated floating point drift. 

\textbf{Optimisation procedure.}\quad
We use the AdamW optimizer \citep{loshchilov2017decoupled} with cosine learning-rate annealing over the full training horizon of \num{20000} optimization steps.
Further optimization methods, we implement are (\emph{i}) parameter updates using exponential moving average, (\emph{ii}) early stopping with a patience of \num{1000} steps (a high number, to not let optimisation difficulties be a bottleneck, but to still save computation), and (\emph{iii}) checkpoint saving.
We use the EMA updates for all sampling and evaluation methods.
We sample mini-batches independently from the source and target distributions. 
For each batch, interpolation times are sampled uniformly $t \sim \mathcal{U}(0, 1)$.
Table~\ref{tab:hyperparameters} gives an overview of other hyperparameters.

\begin{table}[h]
\centering
\caption{Model hyperparameters and sweep configurations.}
\resizebox{\textwidth}{!}{
\begin{tabular}{lccccccc}
\toprule
\textbf{Experiment} & \textbf{Hidden layers} & \textbf{Batch size} & \textbf{Learning rate} & \textbf{$d$} & \textbf{$|\kappa|$} & \textbf{Seeds} \\
\midrule
\textbf{GtG, SGR, MoG}           & 8 & 512 & $1 \times 10^{-4}$ & 2, 3, 4, 8, 16, 32, 64, 256, 512 & 0, 1, 2, 3, 5, 10 & 34, 42, 50 \\
\bottomrule
\end{tabular}
}
\label{tab:hyperparameters}
\end{table}

\textbf{Numerical integration.}\quad
Our implementation generates sampling trajectories by integration of the learned vector field.
We utilise adaptive Dormand-Prince (\texttt{dopri5}) ODE solvers (a form of Runge-Kutta) as a default while having fixed-step Euler as a fallback. 
For geometry-restrictive evaluation experiments, we implement explicit Euler integration with manifold re-projection after every update step for non-local coordinates,
\begin{equation}
    x_{t+\Delta t} = \mathrm{Proj}_{\mathcal{M}_\kappa^d}\!\bigl(x_t + \Delta t\,\boldsymbol{v}_\theta(t, x_t)\bigr),
\end{equation}
while for local coordinates we use $x_{t+\Delta t} = \exp_{x_t}^\kappa\!\bigl(\Delta t\,\boldsymbol{v}_\theta(t, x_t)\bigr)$ as the update.
This projection-oriented integration helps avoid numerical drift away from the manifold when trajectories are long.
Projection here, for spherical manifolds, refers to normalising onto the sphere, while for hyperbolic methods, the points are projected back to the Poincar\'e ball radius (valid region), when the numerical updates reach the boundary.
We also keep the tolerances for the solver and the step sizes fixed across all geometries.

\textbf{Dataset generation.}\quad
Generation of the synthetic datasets is done directly in the tangent space and then mapped to the manifold using the exponential map, see Eq.~\eqref{eq:exp_and_log_map}.
Our framework supports the three experiments introduced in Section~\ref{sec:cross-curvature_experimental_setup} and explained in detail in Appendix~\ref{ap:experimental_setup}.
We have implemented a unified geometry-aware API for data generation to guarantee an identical structure on all the spaces we use.
Training samples can be generated on the fly or reused when the training run requires more samples than the dataset contains.
All partitions (that is: train, validation, test) are generated using deterministic seeded splits, such that we fix all sources of randomness.

\textbf{Deterministic evaluation pipeline.}\quad
To avoid variance in stochastic evaluation, we introduce a deterministic fixed pair evaluation method.
For all experiments we generate source-target samples once, followed by sampling the temporal evaluation probes once.
All subsequent runs reload identical evaluation artifacts. 
The artifacts stored consist of the paired sample, temporal probe grids, interpolation trajectories, predicted vector fields, target vector fields, transported endpoint samples and the metadata of the configuration. 
The configuration metadata is verified for exact compatibility before re-use.
This helps comparison between learned and analytic transport trajectories across runs. 

\textbf{MMD implementation.}\quad
For all MMD evaluations, we use an RBF kernel
$k(x,x')=\exp(-d_{\mathrm{eval}}(x,x')^2/(2\ell^2))$, see Section~\ref{sec:metrics}.
Unless stated otherwise, $\ell$ is chosen by a median heuristic: for each evaluation, we compute the median of the non-zero cross-distances between generated and target samples.
We recompute $\ell$ separately for geodesic, tangent-space, and ambient MMD, since each distance induces a different distance matrix.
For geodesic MMD, we use the same RBF-like statistic with $d_{\mathrm{eval}}=d_{\mathrm{geodesic}}$, but, unlike the ambient and tangent-space RBF kernels, this geodesic kernel is not guaranteed to be positive definite for every manifold configuration.
So we interpret it as an empirical kernel discrepancy.

\textbf{Stability and numerical safeguards.}\quad
To stabilise optimisation and sampling, multiple safeguards are implemented.
We have already talked about tangent-space re-projection and the numerical stability in the manifolds above.
To avoid division by zero in the many manifold operations where quantities may become very small, all denominators have geometry dependent epsilon thresholds, $d \leftarrow \max(d,\epsilon)$.
Here $\epsilon$ is different for \texttt{float32} and \texttt{float64} computations.
Despite this trajectory integration occasionally become unstable due to very large curvature or step size.
So, during training and evaluation trajectories are continuously checked for \texttt{NaN}-values, \texttt{inf}-values, constraint violations (manifold-specific) and exploding tangent norms.

Further to check stability we define a few dignostic statistics which are recorded throughout training and evaluation, these include mean and max of tangent norms, magnitude of the tangent-space violation, frequency of \texttt{NaN}s, failure rate of integration, error of re-projection in a manifold and solver step counts (see also Appendix~\ref{ap:metrics_stability}).
Monitoring these provides additional insights into whether observed outcome is achieved through representational effects or numerical instability. 

\textbf{Reproducibility.}\quad
To enable accurate reproducibility all the experiments use a deterministic seeded initialisation in dataset generation, data partitioning, source-target sampling, parameter initialisation of the model, generation of temporal probe, and subset selection for evaluation.
We keep the validation and test subsets fixed across all runs.
Additionally, we implement artifact-cached evaluation protocol, this helps generate sample and trajectories for every experiment configuration once to re-use of subsequent runs, each artifact bundle is stored together with the configuration metadata.
We run the experiment across three seeds to further verify the reproducibility of the code. 
\clearpage
\section{Additional results}\label{ap:additional_results}
This Appendix supplements Section~\ref{sec:experiments} with additional observations, results, and discussion.
Following the structure of Section~\ref{sec:experiments}, we divide this Appendix into \emph{metric dependence} (\ref{ap:metric_dependence}), \emph{controlled noise perturbations} (\ref{ap:controlled_noise_perturbations}), and \emph{geometric interpretation} (\ref{ap:geometric_interpretation}).
We additionally report numerical stability verifications in Appendix~\ref{ap:numerical_stability}.

\subsection{Metric dependence}\label{ap:metric_dependence}

\begin{figure*}[h]
\captionsetup[subfigure]{
    justification=centering,
    singlelinecheck=false
}
    \centering

    \begin{subfigure}[t]{0.32\textwidth}
        \centering
        \includegraphics[width=\linewidth]{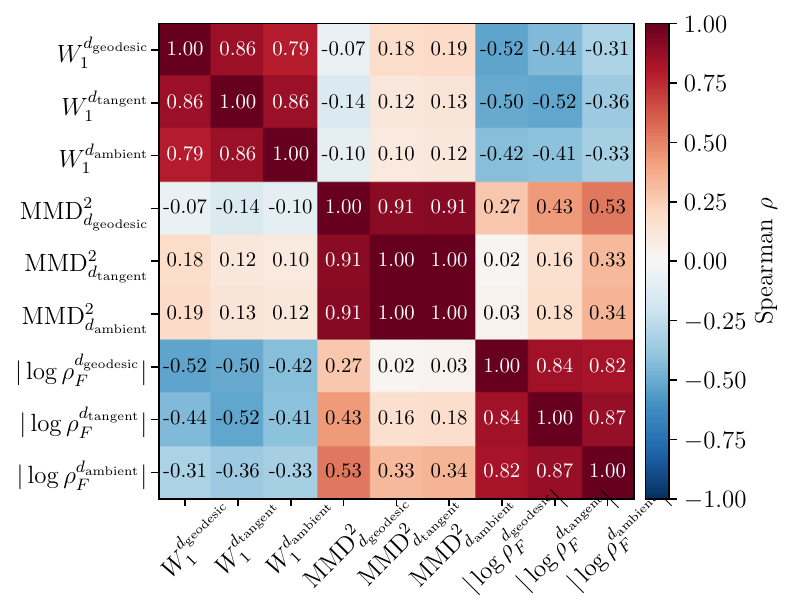}
        \caption{GtG}
        \label{fig:metric_agreement_gtg_app}
    \end{subfigure}
    \hfill
    \begin{subfigure}[t]{0.32\textwidth}
        \centering
        \includegraphics[width=\linewidth]{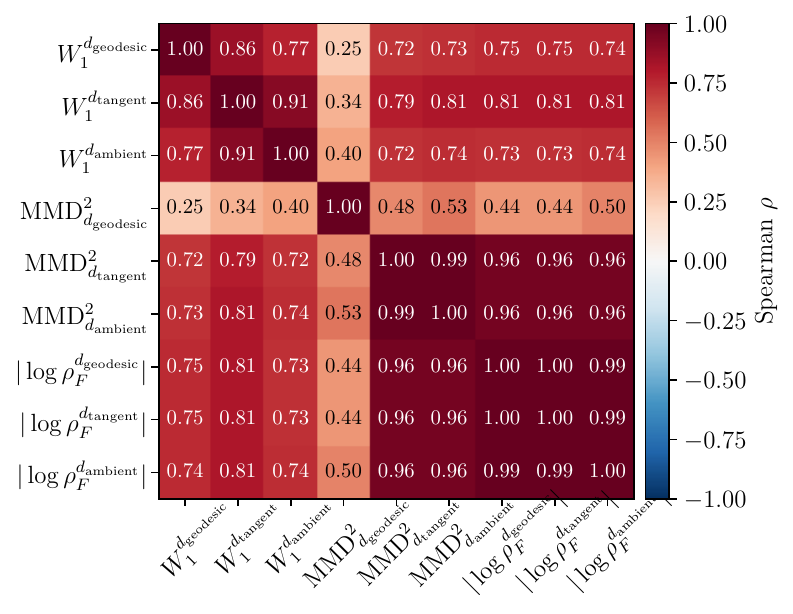}
        \caption{SGR}
        \label{fig:metric_agreement_sgr_app}
    \end{subfigure}
    \hfill
    \begin{subfigure}[t]{0.32\textwidth}
        \centering
        \includegraphics[width=\linewidth]{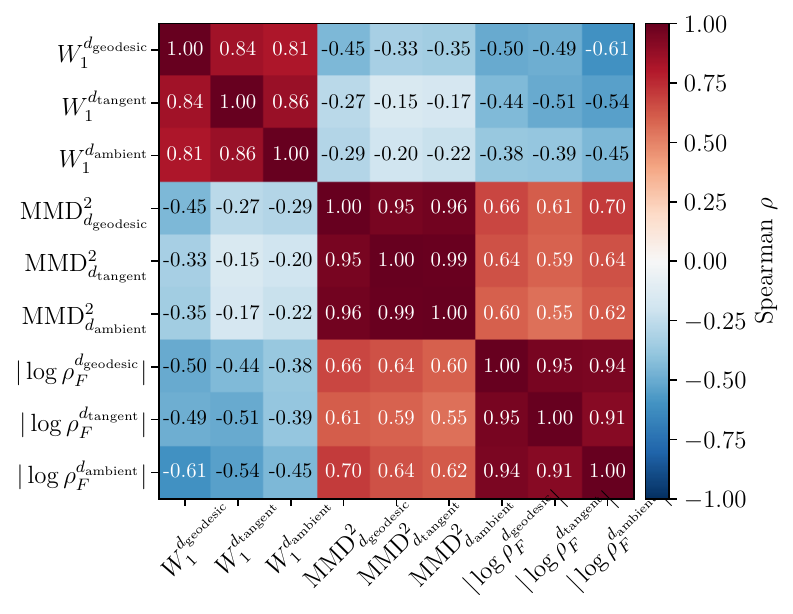}
        \caption{MoG}
        \label{fig:metric_agreement_mog_app}
    \end{subfigure}
    
    \caption{
    \textbf{Metric-type agreement is target-dependent.}
    Spearman correlations are computed across all $(\kappa,d,\mathrm{seed})$-configurations for each target.
    For Fr\'echet variance we use $|\log\rho_F|$ so lower values indicate better quality.
    Distance representations are strongly correlated within a metric type, but agreement across $W_1$, $\mathrm{MMD}^2$, and Fr\'echet variance is weaker and depends on target structure.
    }
    \label{fig:metric_agreement_all_app}
\end{figure*}

Figure~\ref{fig:metric_agreement_all_app} separates two effects: within a metric type, rankings are relatively stable across distance representations; for example, the three $W_1$ variants correlate at $\rho=0.77$ to $0.91$ across targets.
Across metric types, however, agreement is weak and target-dependent: for example, intrinsic $W_1$ and $\mathrm{MMD}^2$ correlate at $-0.07$ on GtG, $0.25$ on SGR, and $-0.45$ on MoG.
Thus, changing from geodesic to tangent or ambient distance often changes scale more than rank, whereas changing metric type can change the conclusion itself.

This difference is expected from the metric definitions.
$W_1$ is transport-based and penalises how far mass must move, whereas $\mathrm{MMD}^2$ compares within- and cross-sample kernel similarities and Fr\'echet variance measures global spread.
These metrics agree most for the rotationally symmetric SGR because the main error is radial: if samples are placed at the wrong radius, transport cost, kernel overlap, and Fr\'echet variance all worsen together.
They can disagree on MoG because matching just the spread is not sufficient; a model can have the right variance while assigning too much or too little mass to individual modes.

\begin{table}[h]
\centering
\captionsetup{
    font=normalfont,
    skip=3pt,
    justification=raggedright,
    singlelinecheck=false
}
\scriptsize
\setlength{\tabcolsep}{1.5pt}
\renewcommand{\arraystretch}{0.85}
\begin{tabular}{lccccc}
\toprule
Curvature & $d=3$ & $16$ & $64$ & $256$ & $512$ \\
\midrule
Hyperbolic & 1.00 & 0.99 & 0.77 & 0.18 & 0.11 \\
Spherical  & 1.00 & 1.01 & 0.95 & 1.01 & 1.01 \\
\bottomrule
\end{tabular}
\caption{\textbf{Ambient Fr\'echet variance underestimates high-$d$ hyperbolic dispersion.}
Entries show $\rho_F^{d_\mathrm{ambient}}/\rho_F^{d_\mathrm{geodesic}}$ for RFM SGR samples.
Values below $1$ mean ambient evaluation reports a more small dispersion error than geodesic evaluation.
The gap grows with dimension in hyperbolic space but stays near $1$ on spheres.
}
\label{tab:sgr_frechet_variance_ratio_dimension_comparison}
\end{table}

\begin{figure}[h]
    \centering
    \includegraphics[width=0.9\linewidth]{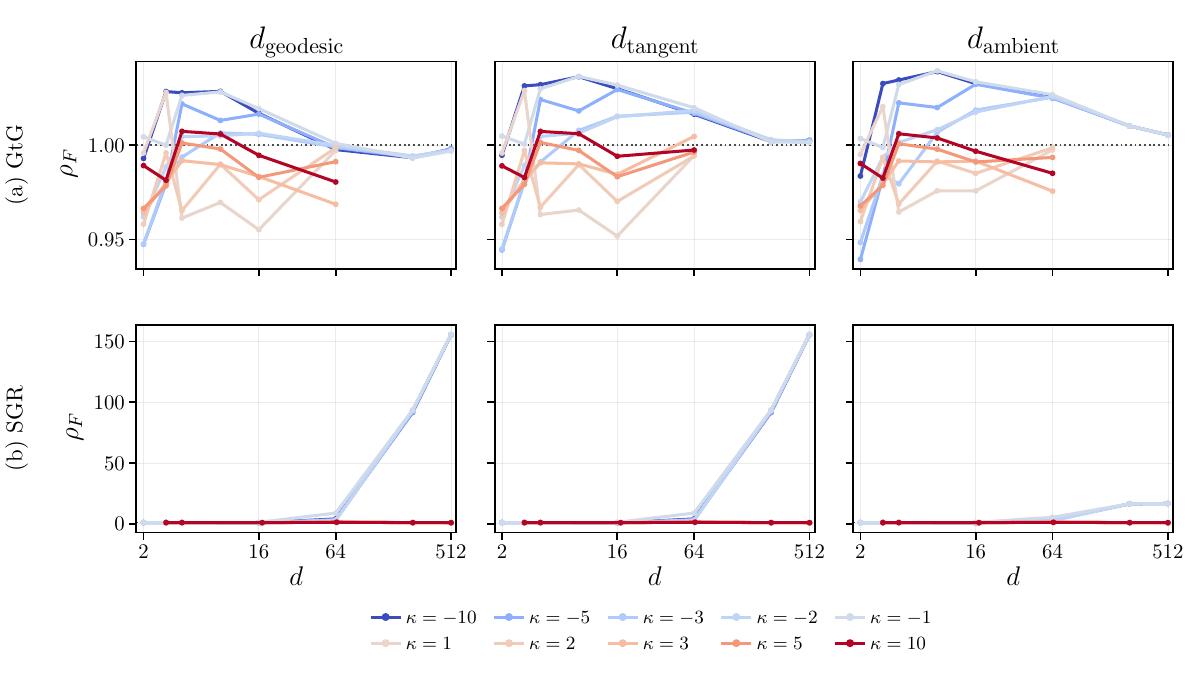}
    \caption{\textbf{Dispersion errors depend on target structure and representation.}
    Ratio of predicted-to-target Fr\'echet variance across signed curvature $\kappa$ and dimension $d$, using geodesic, tangent-space, and ambient distances.
    GtG remains close to $1$ across the sweep, while SGR develops large high-$d$ overdispersion under geodesic and tangent-space evaluation.
    Ambient hyperbolic evaluation compresses this magnitude, consistent with Table~\ref{tab:sgr_frechet_variance_ratio_dimension_comparison}.
    }
    \label{fig:rq2_fr}
\end{figure}

As an additional spread diagnostic, Figure~\ref{fig:rq2_fr} reports the ratio between the generated and target Fr\'echet variances.
A value near $1$ means that the generated samples have approximately the right spread around their Fr\'echet mean.
For GtG, the ratio stays close to $1$, so the generated samples match the target spread well in this baseline case.
SGR, though, shows different behaviour.
At low dimensions, the ratio is also close to $1$, but at high dimensions the geodesic and tangent-space evaluations show high overdispersion in negative-$\kappa$ spaces.
The spherical cases remain much closer to the target spread.
Ambient hyperbolic evaluation reports much smaller errors because Poincar\'e coordinates lie inside a bounded ball; large intrinsic separations are therefore compressed when they are measured with ambient Euclidean distance.
Table~\ref{tab:sgr_frechet_variance_ratio_dimension_comparison} summarises this compression: in hyperbolic space, the ambient-to-geodesic Fr\'echet ratio drops from $0.77$ at $d=64$ to $0.11$ at $d=512$, while it remains close to $1$ on spheres.
Thus, ambient evaluation can still indicate that a model is poor, but it can substantially understate how large the dispersion error is.

\subsection{Controlled noise perturbations}\label{ap:controlled_noise_perturbations}

Table~\ref{tab:rq1_det_scores} checks whether each metric increases consistently with controlled noise.
This is important because our SNR summaries compare metric differences across noise levels; if a metric is not monotone, positive and negative differences can cancel.
$W_1$ is near-monotone across dimensions, while $\mathrm{MMD}^2$ weakens at high $d$, especially at $d=512$.
Thus, high-dimensional MMD SNR values should be interpreted together with the raw response curves.

\begin{table}[h]
\centering
\scriptsize
\setlength{\tabcolsep}{2.2pt}
\renewcommand{\arraystretch}{0.82}
\begin{tabular}{rcc}
\toprule
$d$ & $\rho_{W_1}$ & $\rho_{\mathrm{MMD}}$ \\
\midrule
2   & 1.0000 & 0.9903 \\
3   & 1.0000 & 0.9894 \\
4   & 1.0000 & 0.9940 \\
8   & 1.0000 & 0.9949 \\
16  & 0.9997 & 0.9940 \\
64  & 0.9959 & 0.9760 \\
256 & 1.0000 & 0.8407 \\
512 & 1.0000 & 0.1978 \\
\bottomrule
\end{tabular}
\vspace{0.3em}
\caption{\textbf{Geodesic $W_1$ remains monotone, while geodesic $\mathrm{MMD}^2$ weakens at high $d$.}
Mean Spearman correlations between noise level $\sigma$ and metric value on GtG, averaged over curvatures and seeds.
Values near $1$ indicate that the metric increases consistently with controlled noise. Spherical configurations at $d\in\{256,512\}$ are excluded (see Appendix~\ref{ap:MoG}).
}
\label{tab:rq1_det_scores}
\end{table}

\begin{figure*}[h]
    \centering
    \begin{subfigure}[t]{0.8\textwidth}
        \centering
        \includegraphics[width=\linewidth]{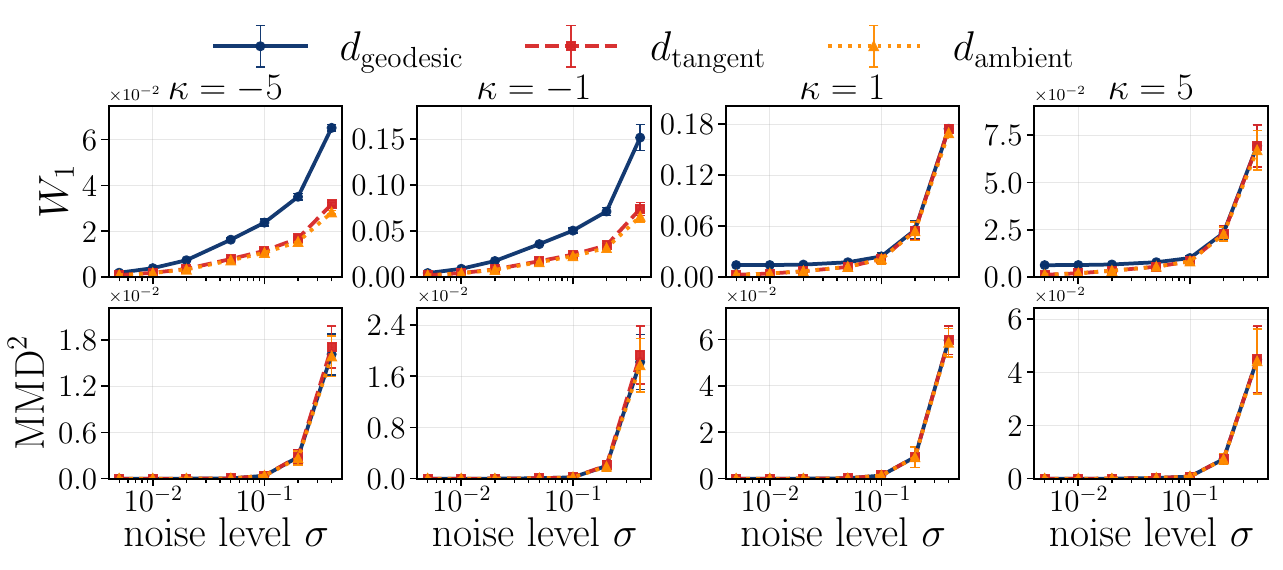}
        \caption{$d=2$}
    \end{subfigure}

    \begin{subfigure}[t]{0.8\textwidth}
        \centering
        \includegraphics[width=\linewidth]{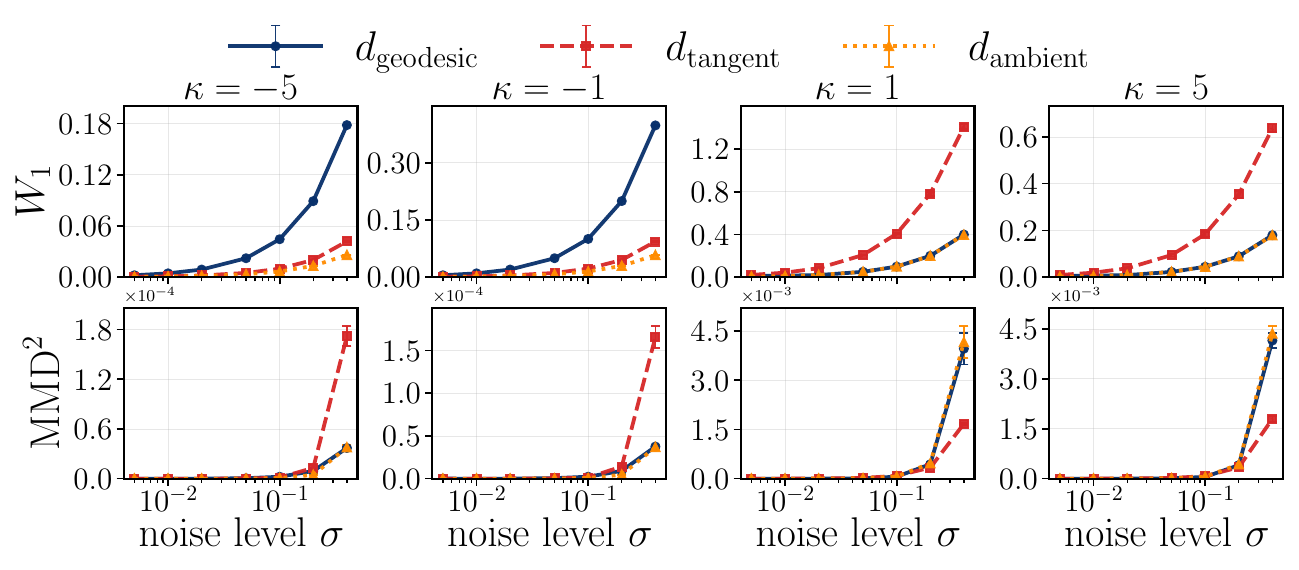}
        \caption{$d=64$}
    \end{subfigure}
    \caption{\textbf{Noise responses depend on dimension and evaluation distance.}
    At $d=2$ and $d=64$, both $W_1$ and $\mathrm{MMD}^2$ generally increase with controlled noise.
    However, the responses differ per evaluation distance, especially in hyperbolic space.
    }
    \label{fig:raw_noise_low_high_dimension_app}
\end{figure*}

\begin{figure*}[h]
    \centering
    \includegraphics[width=0.8\textwidth]{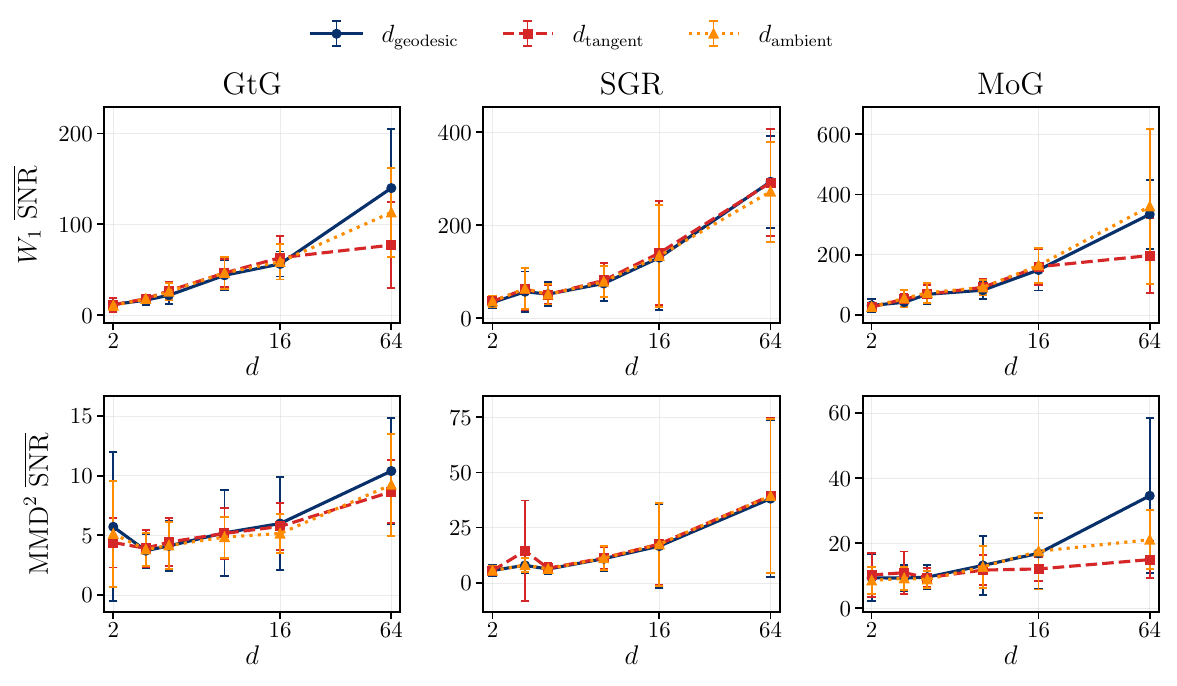}
    \caption{\textbf{Noise-detection reliability varies by metric and target structure.}
    SNR is shown as a function of dimension for GtG, SGR, and MoG, using geodesic, tangent-space, and ambient distances.
    This supports the main-text claim that reliability is not a property of the metric alone.
    }
    \label{fig:absolute_snr_vs_dimension_app}
\end{figure*}

Figures~\ref{fig:raw_noise_low_high_dimension_app} and~\ref{fig:absolute_snr_vs_dimension_app} expand the noise analysis of Section~\ref{sec:experiments}.
The raw SNR curves show both metrics generally increase with noise, but by how much depends on the evaluation distance, especially in hyperbolic space.
The SNR summary also shows that $W_1$ generally becomes more sensitive with dimension, while $\mathrm{MMD}^2$ has a more variable response on the GtG, SGR, and MoG.
Thus, controlled perturbations provide a useful calibration check before trusting a metric for model comparison.

\clearpage
\subsection{Geometric interpretation}\label{ap:geometric_interpretation}

\begin{figure*}[h]
    \centering
    \includegraphics[width=0.68\textwidth]{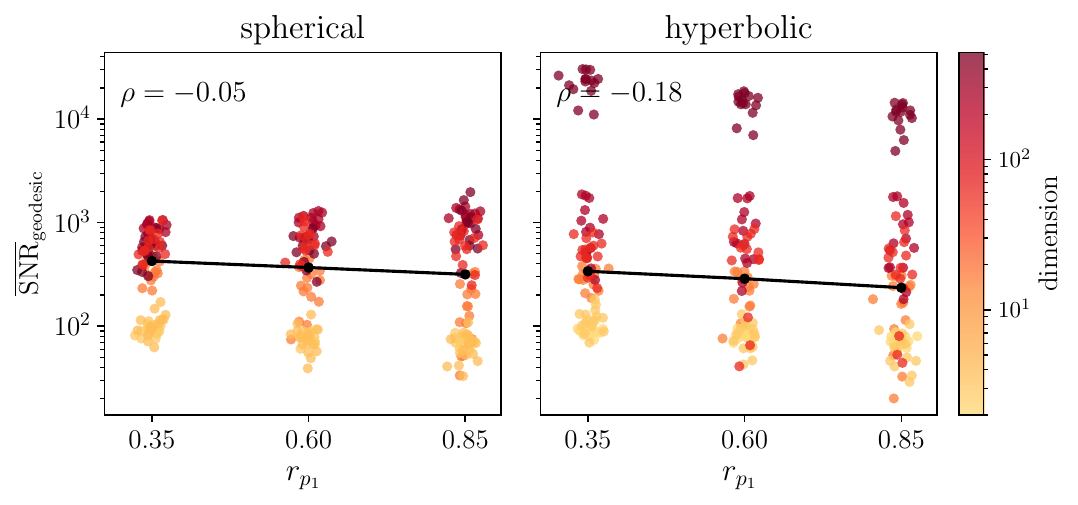}
    \caption{\textbf{SGR radius alone does not determine geodesic SNR.}
    Mean geodesic SNR is shown against the target radius $r_{p_1}$ in spherical and hyperbolic geometries, coloured by dimension.
    SNR remains high across radii, while dimension explains much of the variation.
    }
    \label{fig:snr_geodesic_radius_app}
\end{figure*}

The SGR target was designed to vary radial placement while keeping the angular structure simple by having it close to rotational symmetry.
Figure~\ref{fig:snr_geodesic_radius_app} tests whether this radial parameter alone explains noise-detection reliability.
Mean geodesic SNR remains high across the tested radii, but it shows a mild downward trend as $r_{p_1}$ (also referred to as $r_\mathrm{SGR}$ in Appendix~\ref{ap:Gaussian_ring}) increases, especially in hyperbolic space.
This is consistent with the intuition behind the volume-growth gap (Eq.~\eqref{eq:volume_growth_gap}, defined in Section~\ref{ap:metrics_volume scaling}): moving the target farther from the origin exposes the perturbation to stronger curvature-induced volume distortion.
However, the strongest visual separation is still by dimension rather than by $r_{p_1}$.
Thus, increasing the SGR radius does not by itself explain the reliability gaps; the relevant quantity is how radius interacts with curvature and dimension through $J_\kappa$ (Eq.~\eqref{eq:Jacobian_factor}).

\begin{figure*}[h]
    \centering
    \includegraphics[width=0.72\textwidth]{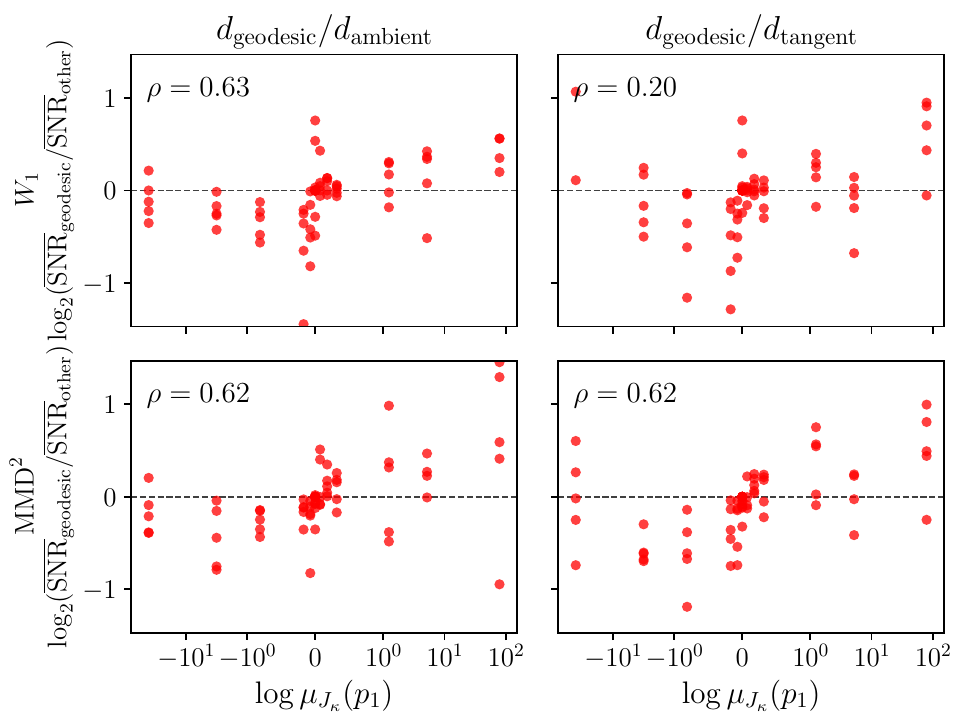}
    \caption{\textbf{Volume scaling explains some, but not all, SNR gaps.}
    Each point is a $(\kappa,d)$-configuration on GtG.
    The x-axis shows target volume scaling, $\log\mu_{J_\kappa}(p_1)$, and the y-axis shows the geodesic SNR advantage over the compared distance.
    Positive values mean geodesic evaluation detects controlled noise more reliably.
    Volume scaling correlates with the geodesic advantage over ambient $W_1$ and over both MMD alternatives, with the relationship weaker for tangent space $W_1$.
    }
    \label{fig:volume_snr_scatter_gtg}
\end{figure*}

Figure~\ref{fig:volume_snr_scatter_gtg} relates reliability gaps directly to the target volume scaling $\log\mu_{J_\kappa}(p_1)$.
For $W_1$, volume scaling is associated with the geodesic advantage over ambient evaluation, but not over tangent space evaluation.
This matches the geometry: ambient distances can compress large intrinsic separations, while the log map keeps radial information more explicit.
For $\mathrm{MMD}^2$, both ambient and tangent alternatives show positive associations, suggesting that volume scaling also changes the pairwise distances used by the RBF kernel.
Still, MMD also depends on bandwidth and can lose sensitivity in high dimensions \citep{ramdas2015decreasing}, so volume scaling is only part of the explanation.

\subsection{Numerical stability checks}\label{ap:numerical_stability}

We check whether the metric differences in the main text could be caused by numerical instability in RFM sampling.
Table~\ref{tab:stability} summarises run-level diagnostics over the same RFM sweeps.
The results show that the observed metric differences are unlikely to be artifacts of failed sampling.
All completed runs produce finite samples, and relative tangency violations remain very small (median $7.1\times10^{-5}$, maximum $5.3\times10^{-4}$).
We therefore interpret the main trends as valid evaluation effects.

\begin{table}[h]
\centering
\small
\caption{\textbf{Numerical stability for RFM sweeps.}
All completed runs produce finite samples, and relative tangency violations remain very small.
This supports the interpretation that the metric differences in the text are not caused by invalid samples or off-manifold vector fields.
The p95 vector-field norm is included as a secondary scale check.}
\begin{tabular}{lrrrccrr}
\toprule
 & & \multicolumn{2}{c}{Sample validity} & \multicolumn{2}{c}{Relative tangency violation} & \multicolumn{2}{c}{$\|v_t(x)\|_{g_\kappa}$ (p95)} \\
\cmidrule(lr){3-4}\cmidrule(lr){5-6}\cmidrule(lr){7-8}
Target & \# runs & finite (\%) & failed (\%) & median & max & median & max \\
\midrule
GtG & 330 & 100.0 & 0.0 & $4.5\times10^{-6}$ & $4.6\times10^{-5}$ & 0.45 & 20.40 \\
SGR & 1350 & 100.0 & 0.0 & $2.3\times10^{-6}$ & $5.3\times10^{-4}$ & 0.71 & 5.20 \\
MoG & 1980 & 100.0 & 0.0 & $1.0\times10^{-5}$ & $2.9\times10^{-4}$ & 0.45 & 12.76 \\
\midrule
All & 3660 & 100.0 & 0.0 & $7.1\times10^{-6}$ & $5.3\times10^{-4}$ & 0.52 & 20.40 \\
\bottomrule
\end{tabular}
\label{tab:stability}
\end{table}

\subsection{Limitations}\label{ap:limitations}
Our experiments are deliberately controlled: we use synthetic targets on constant-curvature manifolds, one RFM implementation, and three random seeds.
Because of this, the SNR values should be interpreted as practical reliability checks, they are not precise uncertainty estimates.
Also, the volume-scaling results point to a likely geometric factor, but do not give a complete causal explanation.
The results further depend on choices such as Sinkhorn regularisation for $W_1$ and the RBF kernel and median-bandwidth rule for $\mathrm{MMD}^2$, so better bandwidth calibration remains future work. 
Further future work should test other types of errors, metric settings, and generative models, as well as real data on more general or learned manifolds.

\end{document}